\documentclass{article}
\usepackage{ijcai21}

\usepackage{times}
\usepackage{soul}
\usepackage{url}
\usepackage[hidelinks]{hyperref}
\usepackage[utf8]{inputenc}
\usepackage[small]{caption}
\usepackage{graphicx,float,subfigure,multirow}
\usepackage{amsmath}
\usepackage{amsthm}
\usepackage{booktabs}
\usepackage{algorithm}
\usepackage{algorithmic}
\usepackage{tikz}
\usepackage[nolist]{acronym}

\makeatletter
\let\NAT@parse\undefined
\makeatother

\usepackage{xcolor}
\colorlet{secondary}{white!25!black}
\usepackage{tabularx, multirow, graphics, url, float, subfigure, amsmath, tikz, hyperref,textcomp}
\hypersetup{
    colorlinks=true,
    citecolor=green!60!black,
    linkcolor=green!60!black,
    urlcolor= cyan!30!black
}

\usetikzlibrary{mindmap,trees,fit,decorations.pathreplacing,patterns,decorations.markings}
\newcounter{mynode}
\tikzset{step node/.code={\stepcounter{mynode}
}}

\tikzstyle{pyschologyBar} = [rounded corners=1mm,fill=cyan, draw=white]
\tikzstyle{systemBBar} = [rounded corners=1mm,fill=blue!45!green, draw=white]
\tikzstyle{riseFPBar} = [rounded corners=1mm,fill=blue!40!white, draw=white]
\tikzstyle{risePPBar} = [rounded corners=1mm,fill=red!40!white, draw=white]
\tikzstyle{firstDBBar} = [rounded corners=1mm,fill=orange, draw=white]
\tikzstyle{firstWKBar} = [rounded corners=1mm,fill=green!70!red, draw=white]
\tikzstyle{riseEBar} = [rounded corners=1mm,fill=yellow!70!black, draw=white]
\tikzstyle{primitiveBar} = [rounded corners=1mm,fill=blue!60!red, draw=white]
\tikzstyle{updatingBar} = [rounded corners=1mm,fill=red!60!black, draw=white]
\tikzstyle{educationText} = [rectangle, above=.1cm, align=center,,scale=1.0]
\tikzstyle{experienceText} = [rectangle, below=.1cm, align=center,,scale=1.00]

\title{Building Affordance Relations for Robotic Agents - A Review}

\author{
Paola Ard{\'o}n$^*$\and
{\`E}ric Pairet\and
Katrin S. Lohan\and\\
Subramanian Ramamoorthy\And
Ronald P. A. Petrick\\
\affiliations
Edinburgh Centre for Robotics \\
University of Edinburgh \& Heriot-Watt University \\
Edinburgh, Scotland, United Kingdom \\
\emails
$^*$paola.ardon@ed.ac.uk
}

\begin{document}

\maketitle

\begin{acronym}[ransac]
  \acro{LbD}{learning by demonstration}
  \acro{RL}{reinforcement learning}
  \acro{IRL}{inverse reinforcement learning}
  \acro{SVM}{support vector machine}
  \acro{DOF}{degrees-of-freedom}
  \acro{CAD}{computer-aided design}
  \acro{ROI}{regions of interest}
  \acro{MCMC}{Markov chain Monte Carlo}
  \acro{ECV}{early cognitive vision}
  \acro{IADL}{instrumental activities of daily living}
  \acro{CDR}{cognitive developmental robotics}
  \acro{2-D}{two-dimensional}
  \acro{3-D}{three-dimensional}
  \acro{RANSAC}{random sample consensus}
  \acro{RGB-D}{red-green-blue depth}
  \acro{IFR}{International Federation of Robotics}
  \acro{CNN}{convolutional neural networks}
  \acro{KB}{knowledge base}
  \acro{MSE}{mean square error}
  \acro{GWR}{geographically weighted regression}
  \acro{PCA}{principal component analysis}
  \acro{CRF}{conditional random fields}
  \acro{ILGA}{integrated learning of grasps and affordances}
  \acro{OACs}{object action complexes}
  \acro{HRI}{human-robot interaction}
  \acro{CP}{control program}
  \acro{AI}{artificial intelligence}
\end{acronym}

\begin{abstract}
 Affordances describe the possibilities for an agent to perform actions with an object. While the significance of the affordance concept has been previously studied from varied perspectives, such as  psychology and cognitive science, these approaches are not always sufficient to enable direct transfer, in the sense of implementations, to  \ac{AI}-based systems and robotics. However, many efforts have been made to pragmatically employ the concept of affordances, as it represents great potential for AI agents to effectively bridge perception to action.
 In this survey, we review and find common ground amongst different strategies that use the concept of affordances within robotic tasks, and build on these methods to provide guidance for including affordances as a mechanism to improve autonomy. To this end, we outline common design choices for building representations of affordance relations, and their implications on the generalisation capabilities of an agent when facing previously unseen scenarios. Finally, we identify and discuss a range of interesting research directions involving affordances that have the potential to improve the capabilities of an AI agent.

\end{abstract}

\section{Introduction \label{sc:introduction}}

    The psychologist James J. Gibson coined the term \textit{affordance} as the ability of an agent to perform a certain action with an object in a given environment \cite{gibson1966senses}. In general, the concept contributes  an encapsulated description of the different ways to comprehend the world \cite{hammond2010affordance}. Nonetheless, \cite{gibson1966senses}'s understanding of affordances generated controversy among psychologists, resulting in a vast diversity of definitions \cite{norman1988psychology,mcgrenere2000affordances}. 
    
    In \ac{AI}, affordances play a key role as intermediaries that organise the diversity of possible perceptions into tractable representations that can support reasoning processes to improve the generalisation of tasks. A number of examples of the use of affordances as a form of inductive bias for learning mechanisms can also be found in robotics. In this regard, robotics is a key frontier area for \ac{AI} that allows for experimental and practical implementations of affordances, with applications to tasks such as action prediction, navigation and manipulation.

    The idea of affordances has been studied from different perspectives. Early surveys \cite{chemero2007gibsonian,csahin2007afford,horton2012affordances} summarise formalisms that attempt to bridge the controversial concept of affordances in psychology with mathematical representations. Other surveys discuss the connection of robotic affordances with other disciplines \cite{jamone2016affordances}, and propose classification schemes to review and categorise the related literature \cite{min2016affordance,zech2017computational}. In contrast, we focus  more on the implications of different design decisions regarding task abstraction and learning techniques that could scale up in physical domains, to address the need for generalisation in the AI sense. As a result, we attempt to capture and discuss the relationship between the requirements, implications and limitations of  affordances in an intelligent artificial agent.
    
    
    
    \begin{figure*}[th!]
        \centering
         
        \subfigure[\protect\cite{kruger2011object} ]{\label{fig:kruger_d} \includegraphics[width=3cm, trim = 0cm 0cm 0cm 0cm]{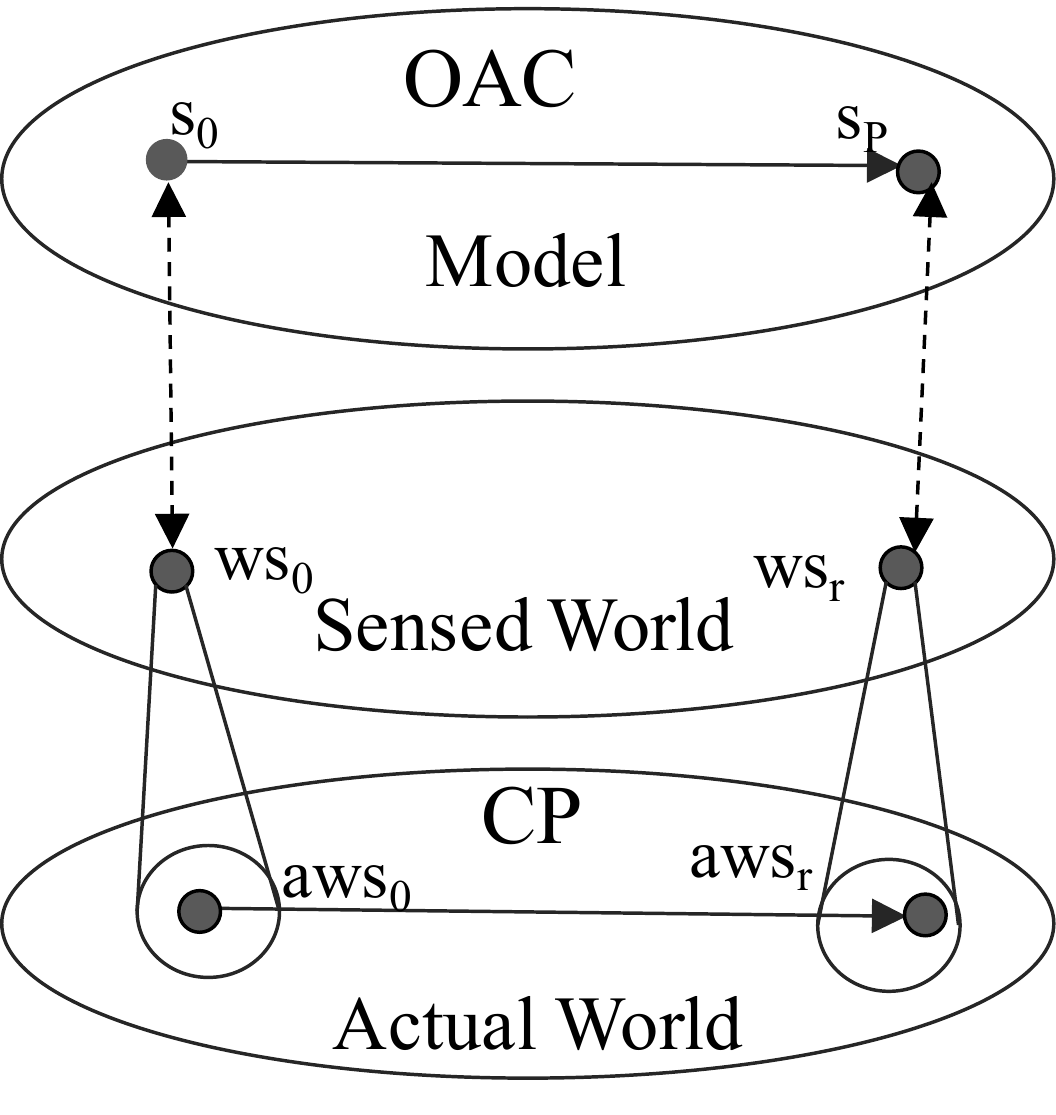}}
        \subfigure[\protect\cite{montesano2007modeling} ]{\label{fig:montesano_d} \includegraphics[width=4cm,trim = 0cm 0cm 0cm 0cm]{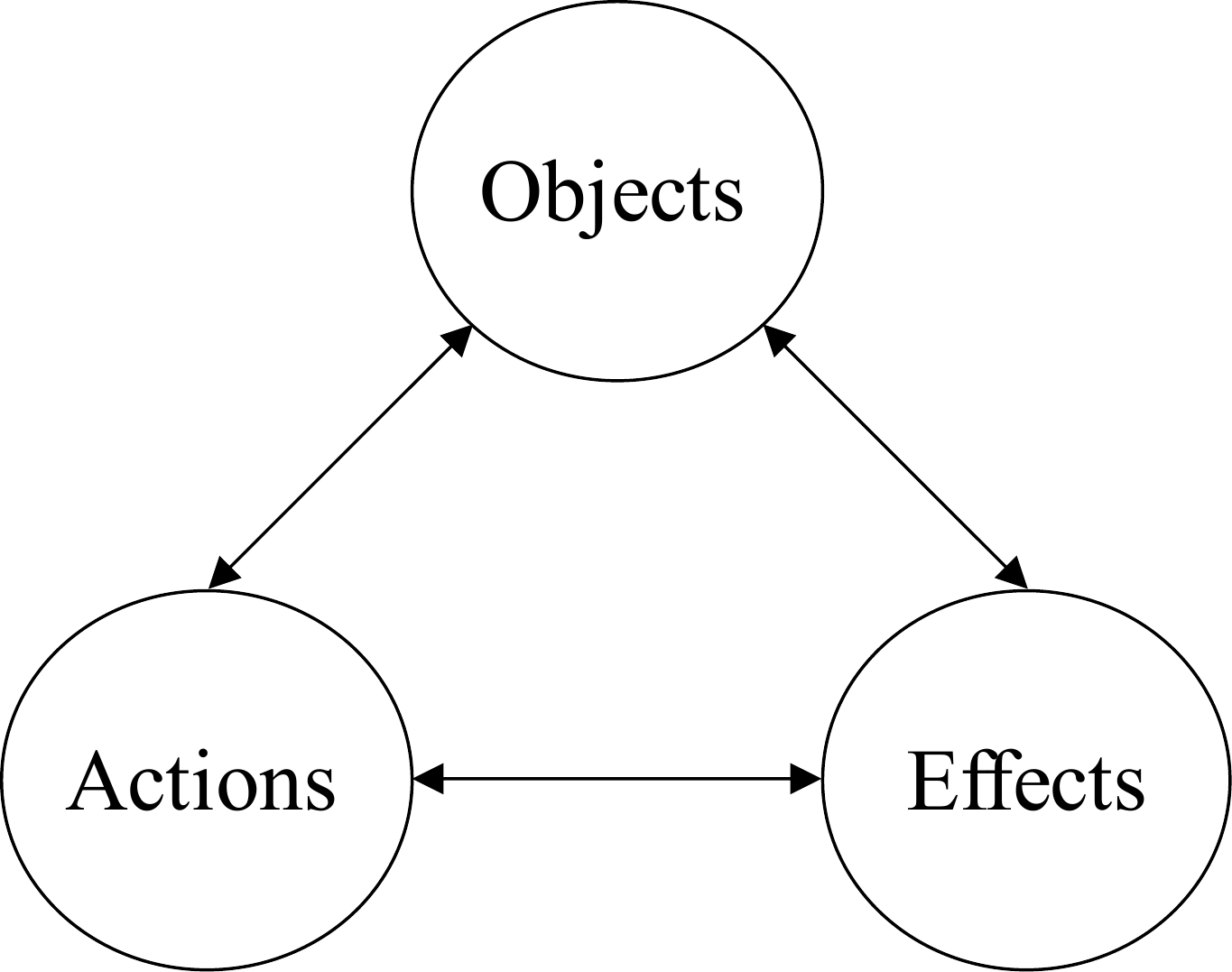}}
        \subfigure[\protect\cite{cruz2016training} ]{\label{fig:cruz_d} \includegraphics[width=3.3cm,trim = 0cm 0cm 0cm 0cm]{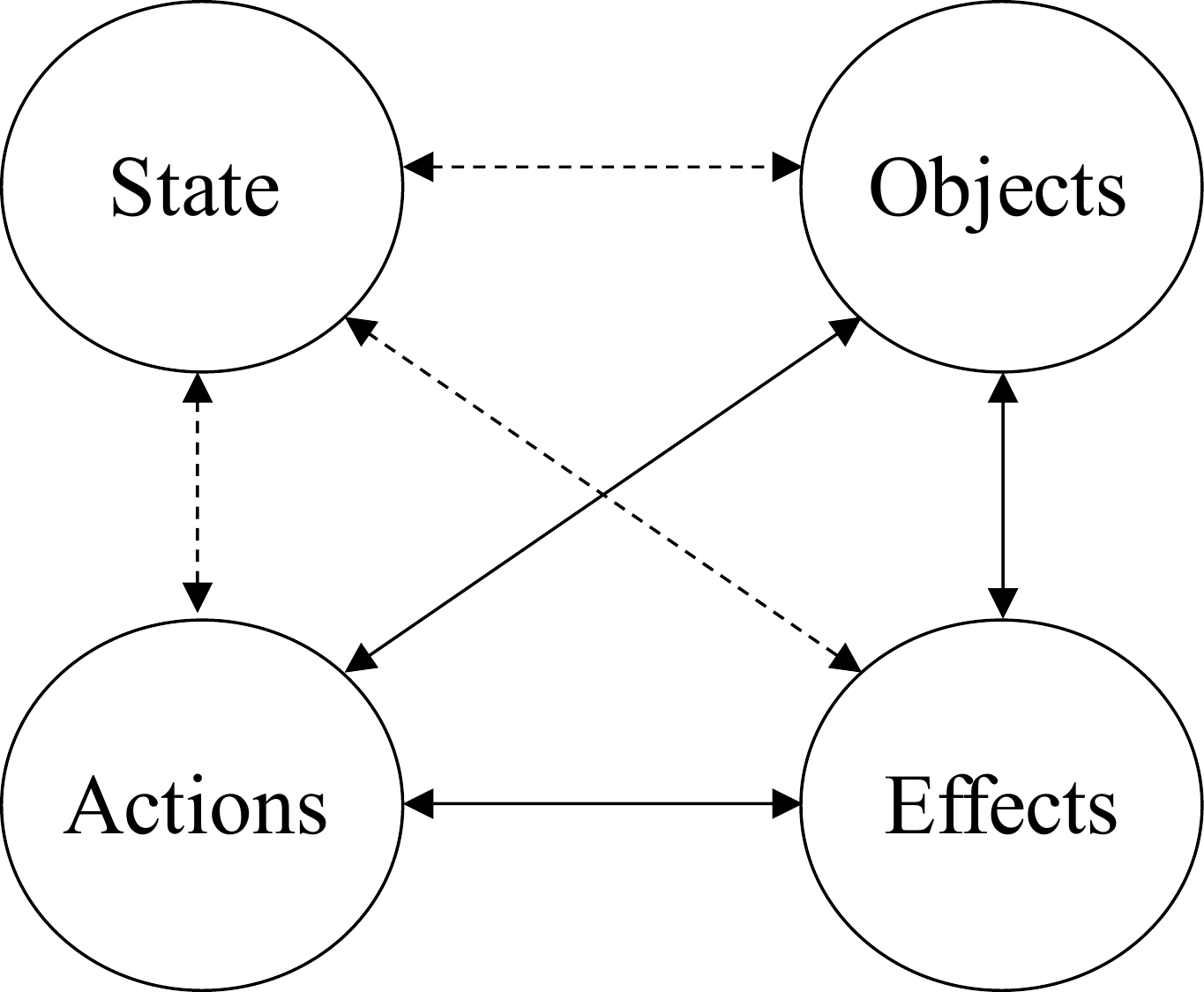}}
        \subfigure[\protect\cite{barck2009learning} ]{\label{fig:barck_d} \includegraphics[width=6cm,trim = 0cm 0cm 0cm 0cm]{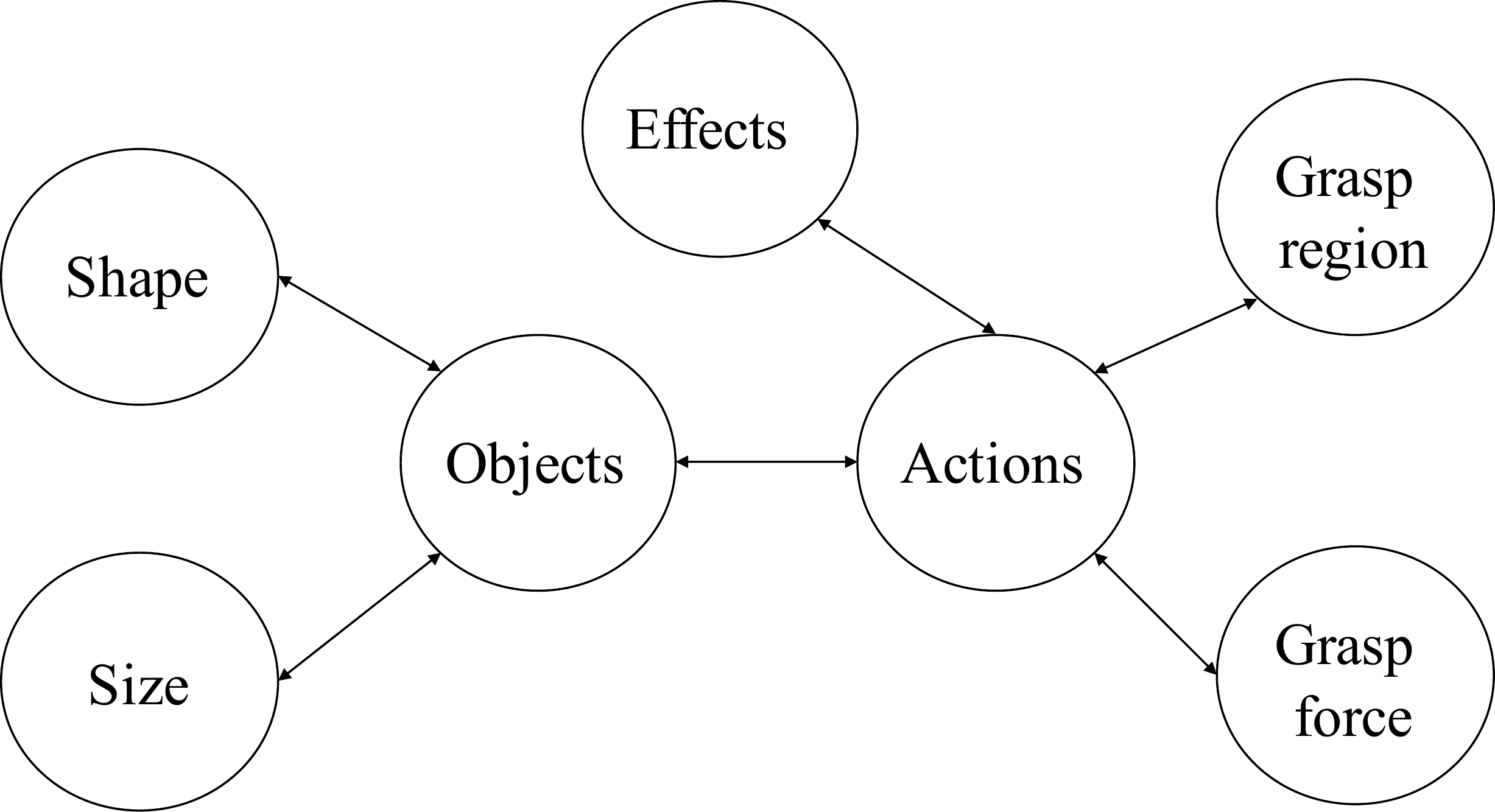}} 
        
        \caption{Approaches that propose a mathematical formalism of the affordance relation. a) \protect\cite{kruger2011object} propose \acf{OACs} as the relationship between the sensed and actual world ($s_0$ is the initial state and $s_p$ is the predicted state). b) \protect\cite{montesano2007modeling} represent affordances as a relation between objects, actions, and effects. Objects are entities which the agent is able to interact with, actions represent the behaviour that can be performed with the object, and effects are the results caused by applying an action. c) \protect\cite{cruz2016training} represent the relations between state, objects, actions and effects, where the state is the current agent's condition and different effects could be produced for different situations. d) \protect\cite{barck2009learning} present an ontology affordance formalism for grasping, composed of three basic elements alongside object properties and grasping criteria.
        }
        \label{fig:formalisms}
        \vspace{-0.3cm}
    \end{figure*}
    The goal of this paper is to provide guidance to researchers wanting to use the concept of affordances to improve generalisation in robotic tasks. As such, we focus on two key questions: \textit{what aspects should be considered when including affordances as a bias for learning and policy synthesis in AI agents?}~and \textit{how does the combination of such aspects influence the generalisation capabilities of the system?} 
    After a thorough literature analysis in Section~\ref{sc:formalism}, we discuss how, regardless of the underlying abstraction of the concept, using affordances usually refers to the problem of perceiving a \textit{target object}, identifying what \textit{action} is feasible with it and the \textit{effect} of applying these actions on task performance. 
    The relation of these three elements from now on will be referred to as the \textit{affordance relation}. The diversity of techniques and their implications when building an affordance relation forms the backbone of this survey.
    As such, our contribution is twofold. First, given that we have identified common aspects that define the affordances concept from the point of view of an \ac{AI} system, we outline the different types of data, processing, learning and evaluation techniques as found in the literature. Moreover, we identify the different levels of \textit{a priori} knowledge on the affordance task. We find that this knowledge directly influences the generalisation capabilities of the system, i.e., the ability to broadly apply some knowledge by inferring from specific cases\footnote{The literature related to this survey is available at \mbox{\url{https://paolaardon.github.io/affordance_in_robotic_tasks_survey/}}\label{url} It includes 152 papers from 2003 to 2020 in an interactive format.}.


\section{Synopsis of Affordance Formalisms \label{sc:formalism}}

    This section summarises the evolution of formalisms that use the concept of affordances to improve an agent's performance. This evolution can be divided into two main stages. The first stage is characterised by mathematical conceptualisations as extensions from psychology theories. The second stage corresponds to formalisms that focus on the capabilities of the system rather than on recreations from psychology.  
    
    \subsection{Psychology-centric formalisms \label{sec:early_formalisms}}
        
        In the early stages of the field, affordances were formalised from different perspectives. Namely, this work emphasised \textit{where} the affordance resided following psychology theories. \cite{chemero2007gibsonian,csahin2007afford} extensively review and discuss existing approaches (up to 2007) that translate psychology perspectives into the robotics field. \cite{csahin2007afford} classified the early affordance literature into three different parallel perspectives: 
        \begin{itemize}
        \itemsep-0.2em 
          \item Agent perspective: The affordance resides inside the agent's possibilities to interact with the environment.
          \item Environmental perspective: The affordance includes the perceived and hidden affordances in the environment.
          \item Observer perspective: The affordance relation is observed by a third party to learn these affordances.
        \end{itemize}
        
The environmental perspective is the most abstract of the perspectives. However, more practically, given the nature of hidden affordances, current methods in applications such as robotics consider either the agent or the observer perspective to build tractable representations of affordance models. 

    \subsection{Agent-centric formalisms \label{sec:current_formalims}}
    
        \cite{csahin2007afford}, besides reviewing physchology-inspired work, proposed the first formalism focused on the agent perspective. They argued that an affordance relation between \textit{effect} and an \textit{(entity, behaviour)} tuple should be considered. As such, when the agent applies the behaviour on the \textit{entity}, the \textit{effect} is generated: (\textit{effect}, (\textit{entity, behaviour})).
        From the agent's perspective, \cite{montesano2007modeling,kruger2011object} define affordances as using symbolic representations obtained from sensory-motor experiences (see Fig~\ref{fig:montesano_d} and Fig~\ref{fig:kruger_d}, respectively). Similarly, \cite{cruz2016training} consider that if an affordance exists and the agent has knowledge and awareness of it, the agent can choose to utilise it given the current state (see Fig.~\ref{fig:cruz_d}).  \cite{barck2009learning} compare the reasoning engine used for learning in the ontological approach in contrast to the voting probabilistic function to examine the generalisation capabilities of the system (see Fig.~\ref{fig:barck_d}).
        
        As observed in this section, the perspectives on how affordance should be included in a system vary significantly. Nonetheless, there are common aspects that can be identified across all formalisms that help us ground the basic requirements of affordances for a given task.

     \subsection{Formalism influence on review criteria \label{sec:selection_criteria}}
     
     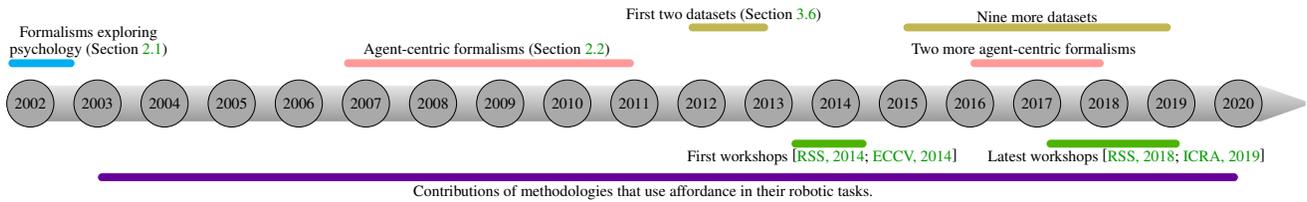
\begin{figure*}[h!]
        \centering
        \resizebox{2.03\columnwidth}{!}{
            \begin{tikzpicture}
            \shade[rounded corners=0cm, top color=gray!20,bottom color=gray!80] (0,5) -- (0,5.80) -- (27.5,5.80) -- (28.5,5.45) -- (28.5,5.35) -- (27.5,5.00);
            
            \draw (0,5.40) node[circle,fill=secondary!45!white,draw]{2002};
            \draw (1.5,5.40) node[circle,fill=secondary!45!white,draw]{2003};
            \draw (3,5.40) node[circle,fill=secondary!45!white,draw]{2004};
            \draw (4.5,5.40) node[circle,fill=secondary!45!white,draw]{2005};
            \draw (6,5.40) node[circle,fill=secondary!45!white,draw]{2006};
            \draw (7.5,5.40) node[circle,fill=secondary!45!white,draw]{2007};
            \draw (9,5.40) node[circle,fill=secondary!45!white,draw]{2008};
            \draw (10.5,5.40) node[circle,fill=secondary!45!white,draw]{2009};
            \draw (12,5.40) node[circle,fill=secondary!45!white,draw]{2010};
            \draw (13.5,5.40) node[circle,fill=secondary!45!white,draw]{2011};
            \draw (15,5.40) node[circle,fill=secondary!45!white,draw]{2012};
            \draw (16.5,5.40) node[circle,fill=secondary!45!white,draw]{2013};
            \draw (18,5.40) node[circle,fill=secondary!45!white,draw]{2014};
            \draw (19.5,5.40) node[circle,fill=secondary!45!white,draw]{2015};
            \draw (21,5.40) node[circle,fill=secondary!45!white,draw]{2016};
            \draw (22.5,5.40) node[circle,fill=secondary!45!white,draw]{2017};
            \draw (24,5.40) node[circle,fill=secondary!45!white,draw]{2018};
            \draw (25.5,5.40) node[circle,fill=secondary!45!white,draw]{2019};
            \draw (27,5.40) node[circle,fill=secondary!45!white,draw]{2020};

            \filldraw[pyschologyBar] (-0.5,6.20) rectangle (1,6.40);
            \node[educationText] at (1.3,6.20){Formalisms exploring \\psychology (Section~\ref{sec:early_formalisms})};
            
            \filldraw[risePPBar] (7,6.20) rectangle (13.5,6.40);
            \node[educationText] at (10.2,6.20){Agent-centric formalisms (Section~\ref{sec:current_formalims})};
            \filldraw[risePPBar] (21,6.20) rectangle (24,6.40);
            \node[educationText] at (22.2,6.20){Two more agent-centric formalisms};
            
            \filldraw[riseEBar] (14.7,7) rectangle (16.5,7.20);
            \node[educationText] at (15.5,7){First two datasets (Section~\ref{sec:available_datasets})};
            \filldraw[riseEBar] (19.5,7) rectangle (25.5,7.20);
            \node[educationText] at (22.5,7){Nine more datasets};
            
            \filldraw[primitiveBar] (1.5,3.65) rectangle (27,3.85);
            \node[experienceText] at (13.7,3.80){Contributions of methodologies that use affordance in their robotic tasks.};
            
            \filldraw[firstWKBar] (17,4.40) rectangle (18.70,4.60);
            \node[experienceText] at (17.70,4.60)%
            {First workshops \cite{workshop1,workshop2}};
            
             \filldraw[firstWKBar] (22.7,4.40) rectangle (25.70,4.60);
            \node[experienceText] at (24.5,4.60)%
            {Latest workshops \cite{workshop3,workshop4}};
        \end{tikzpicture}
        }
      \caption{Timeline of significant events in the field of affordances for robotics tasks. Some of the highlights include the popularity of formalisms that focused on translating psychology theories to robotics, and those that proposed formalisms based solely on what the agent can perceive and do with the objects. Other important events are the release of datasets and events aimed at
      standardising the field. Contributions using affordances in robotics appear from 2003 onwards.
     }
          \label{fig:timeline}
          \vspace{-0.3cm}
     \end{figure*}
     
        In spite of the differences in approaching the problem, for both psychology and agent-centric formalisms, the purpose remains to achieve high levels of generalisation performance. Interestingly, work across both areas builds the affordance relation using the same three elements: a target object, an action to be applied to that target object and an effect that this action produces. Across formalisms, there are equivalences in both terminology (i.e., action $\leftrightarrow$ behaviour) and type of data (i.e., semantic labels, features).
        Especially from agent-centric formalisms, it is notable that depending on the nature of the data, the time when this data is processed and learned affects the agent's generalisation capabilities. We identify that, as in the formalisms, the literature on affordances for artificial agents follows the same three elements to build the affordance relation. Moreover, variations in the processing and learning of the data (as detailed in Section~\ref{sc:design_choices}) constitute different levels of prior affordance relation knowledge, which closely influences the generalisation performance of the agent. Section~\ref{sc:deployment} details this correlation.
        
        Fig.~\ref{fig:timeline} summarises a timeline of important stages in the field, as extracted from reviewing the literature. From the timeline, we can see that including affordances in robotic tasks is a relatively new field. The agent-centric formalisms on which most of the methodologies base their approaches were not created until after $2007$. Moreover, there are still recent proposals trying to ground the view of affordances in robotics. Regarding open source material and workshops, currently there are $11$ online available datasets and there have been four publicly available workshops to discuss affordances in robotics \cite{workshop1,workshop2,workshop3,workshop4}. Gatherings of researchers with similar interests open doors to discussion, and thus, create opportunities for grounding and advancing progress in the field. Expanded and detailed discussions on related issues are presented in Section~\ref{sc:limitations} and \ref{sc:openquestions}.

\section{Design Choices \label{sc:design_choices}}
    
    \begin{figure}[bt!]
            \hypersetup{linkcolor=black}
    
            \centering
            \resizebox{0.9\columnwidth}{!}{
               
              \begin{tikzpicture}
                	\tikzstyle{every node}=[font=\large]
                    \path[mindmap,concept color=gray!80!white,text=black]
                    node[concept,minimum size=4cm] {\textbf{Affordance relation \\ design choices}}
                    [clockwise from=0]                    
                    
                    child[concept color=gray!65!white, level distance=4.2cm,concept color=orange!85!black, style={sibling angle=0}] {
                      node[concept,minimum size=2.5cm] {Learning relations \\ {\small Section~\ref{sec:relational_model_choices}}}
                      [clockwise from=120]
                      child [level distance=3cm] { node[concept,minimum size=2.3cm] {Deter-ministic} }
                      child [level distance=3cm] { node[concept,minimum size=2.3cm] {Probabi-listic} }
                      child [level distance=3cm] { node[concept,minimum size=2.3cm] {Heuristics} }
                      child [level distance=3cm] { node[concept,minimum size=2.3cm] {Planning} }
                    }    
                     child[concept color=gray!65!white, level distance=4.2cm,concept color= blue!45!white, style={sibling angle=-85}] {
                      node[concept,minimum size=2.5cm] {Evaluation \\ {\small Section~\ref{sec:evaluation_choices}}}
                      [clockwise from=100]
                      child[level distance=3cm] { node[concept,minimum size=2.3cm] {\!\!\!Qualitative }}
                      child[level distance=3cm] { node[concept,minimum size=2.3cm] {\!\!\!Quantitative}}
                      }
                    child[concept color=gray!65!white, level distance=4.4cm,concept color=cyan, style={sibling angle=100}] {
                      node[concept,minimum size=2.5cm] {Sensory input \\ {\small Section~\ref{sec:sensory_choices}}}
                      [clockwise from=230]
                      child [level distance=3cm] { node[concept,minimum size=2.3cm]  {Proprio-ception} }
                	  child [level distance=3cm] { node[concept,minimum size=2.3cm] {Kinaes-thetic} }
                      child [level distance=3cm] { node[concept,minimum size=2.3cm]  {Tactile} }
                      child [level distance=3cm] { node[concept,minimum size=2.3cm]  {Visual} }
                    }
                      child[concept color=gray!65!white, level distance=4.5cm,concept color=green!75!black, style={sibling angle=165}] {
                      node[concept,minimum size=2.5cm] {Acquisition techniques \\ {\small Section~\ref{sec:collection_choices}}}
                      [clockwise from=300]
                      child [level distance=3cm] { node[concept,minimum size=2.3cm]  {Image Labels} }
                      child [level distance=3cm] { node[concept,minimum size=2.3cm]  {Demons-trations} }
                      child [level distance=3cm] { node[concept,minimum size=2.3cm]  {\!\!\!Exploration} }
                    }
                    child[concept color=gray!65!white, level distance=3.97cm,concept color= purple!55!white, style={sibling angle=195}] {
                      node[concept,minimum size=2.35cm] {Actions \\ {\small Section~\ref{sec:action_choices}}}
                      [clockwise from=320]
                      child [level distance=2.68cm] { node[concept,minimum size=2.2cm] {Primitve} }
                      child [level distance=2.68cm] { node[concept,minimum size=2.2cm] {\!\!Compound} }
                    }
                    ;
            \end{tikzpicture}
           }
            \caption{Design choices to build affordance relations as considered in this survey. Starting at the top-left branch and counter-clockwise, we outline the options found in the literature for collecting data in terms of sensory input (blue) and acquisition techniques (green). We also look at the type of actions used to build an affordance relation and perform a task (rose). Once the type of affordance elements have been chosen, we outline the different learning methods employed to relate those elements (orange). Finally, we review the different metrics for evaluation of the affordance task (purple).
            }
            \label{fig:design choices}
            \vspace{-0.5cm}
    \end{figure}
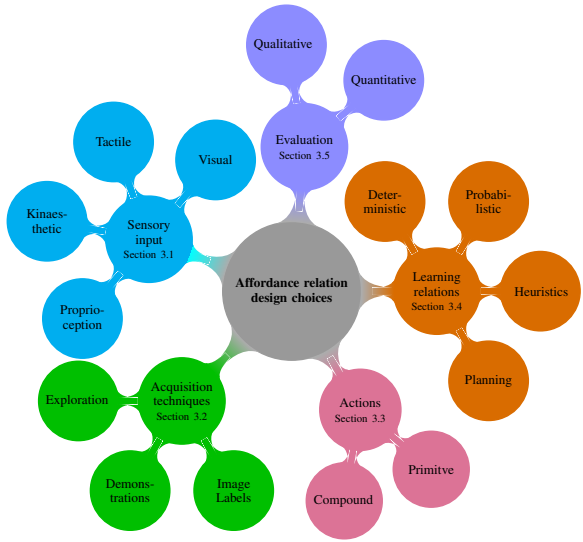
    
    We consider two aspects that influence the generalisation capabilities of an agent when using affordances: the acquisition and processing of the three elements composing an affordance relation model (i.e., target object, actions and effects), which is considered in this section; and
    the level of \textit{a priori} knowledge of the affordance relation, with respect to the task start time, which is  detailed in Section~\ref{sc:deployment}.

    \subsection{Sensory input \label{sec:sensory_choices}} 
        We start by considering the data acquisition \textit{sensory input}, which refers to the medium used to recognise all the physical and visual qualities that suggest a set of actions in the scene. For example, a ball contains the visual and physical features that suggest the affordance to \texttt{roll}. The features can be perceived with a different set of sensors. An extensive summary of perception interaction is presented in \cite{bohg2017interactive}. Common practices in robotic affordance tasks include using \textit{visual} input \cite{zhu2014reasoning,saxena2014robobrain}, \textit{tactile} sensors \cite{baleia2015exploiting}, \textit{kinaesthetic} \cite{katz2014perceiving} and \textit{proprioceptive} sensory feedback \cite{bonaiuto2015learning,kaiser2015validation}. The choice of sensory input is greatly influenced by the purpose of the target task, for example, using visual sensors for recognition tasks \cite{szedmak2014knowledge,myers2015affordance} or a combination of sensory inputs to understand the environment \cite{ardon2020self,bekiroglu2013predicting,bierbaum2009grasp,baleia2015exploiting,liu2019learning,veres2020incorporating}. 

    \subsection{Data collection \label{sec:collection_choices}} 
         Together with the technique for collecting data, one usually also decides on the underlying data structure. For example, consider the task of making an object roll where there are two objects, a ball and an apple, both of which afford \texttt{roll}. The agent could use class labels that detect \texttt{ball} and \texttt{apple} or could identify features that make an object roll, in which case the agent could potentially generalise this behaviour to other objects. From the reviewed literature, the most common practice is to annotate pixel \textit{labels}, through supervised or self-supervised methods, on RGB and RGB-D visual input that represent affordance features \cite{do2018affordancenet,chu2016learning,dehban2016denoising,saxena2014robobrain,stoytchev2008learning,sun2010learning,sinapov2007learning}. Other work uses a combination of visual input with \textit{demonstrations} from a tutor \cite{chu2016learning} and \textit{exploration} techniques \cite{gonccalves2014learning,antunes2016human,wang2013robot,cruz2016training,saxena2014robobrain} to collect the affordance relation.

    \subsection{Deployed actions \label{sec:action_choices}} 
        Tasks an agent might perform 
        could range from pure affordance recognition on a target object to the deployment of an action to achieve a manipulation or navigation task.
        For example, an agent might recognise that a ball affords \texttt{roll} and then apply an action like  \texttt{push} so that the ball rolls. A more complex action might be the compound task of \texttt{handing over} the ball, which would require the agent to \texttt{reach} and \texttt{grasp} the ball and then \texttt{approach} another agent. \cite{asada2009cognitive} summarises work that imitates the cognitive development of an infant, dividing it into $12$ stages according to the difficulty of the motions. Inspired by \cite{asada2009cognitive} in this survey, we group  actions in two sets: \textit{primitive} actions, defined by simple motions, such as turning, moving forward, grasping or pushing \cite{abelha2016model,baleia2015exploiting,seker2019deep}, and \textit{compound} actions defined by combining multiple simple motions, such as pouring or handing over an object \cite{price2016affordance,sun2014object,zhu2014reasoning}. 
       
    \subsection{Learning affordance relations\label{sec:relational_model_choices}}
    
        In addition to acquiring data to associate the affordance elements, it is important to consider the learning model that  encapsulates the affordance relation. We identify four affordance relation learning strategies. In the rolling object example, one might have collected a sample showing that a ball affords \texttt{roll}.
        Given that the relation was learned with one specific rolling example, the agent knows one particular way of rolling. This \textit{deterministic} approach results in a model without any randomness \cite{szedmak2014knowledge}.
        If the collected data could instead contain overlapping examples, including `randomness', then a \textit{probabilistic} learning approach could be applied \cite{zhu2014reasoning,saxena2014robobrain,song2015task,mar2015multi,nguyen2017object,moldovan2012learning}. Another option is for the agent to have a prior set of logical rules that determines a round object in motion on a surface rolls, so it builds the affordance relation model based on \textit{heuristics} \cite{baleia2015exploiting,ugur2011goal,koppula2014physically}. A slightly more difficult scenario would be to make the ball roll and when it reaches a stationary state to make it roll back. Here the relation could be built as a task \textit{planning} problem \cite{aksoy2015model,ugur2011goal,cutsuridis2013cognitive}.
    
    \subsection{Metrics and evaluation\label{sec:evaluation_choices}}
        Across the literature, different approaches might evaluate the same task differently. For instance, in the rolling ball example, an approach might say a task succeeds if the ball rolls, and fails if it does not move. In this case, the method determines if the task completed successfully based on \textit{qualitative} metrics \cite{ardon2019learning,bonaiuto2015learning,cai2019metagrasp}. Another option might be to measure the success of rolling a ball by matching trajectory accuracy or measuring displacement on a surface. Work that uses a numeric-based metric \textit{quantitatively} evaluates the task. Particularly for work using \textit{quantitative} metrics, the evaluation is closely correlated to the application and the purpose of the task. To date, there exists little direct comparison across different applications (i.e., recognition, manipulation, navigation). 
        However, popular metrics in the field include confusion matrices \cite{zhu2014reasoning,ye2017can,song2015task,aksoy2015model}, \ac{MSE} \cite{katz2014perceiving}, and accuracy of classification metrics that reflect intrinsic assessments \cite{myers2015affordance,gonccalves2014learning,aldoma2012supervised,cruz2016training}. Further discussion on the need for standard setups is presented in Sections~\ref{sc:limitations}.
  
    \subsection{Datasets \label{sec:available_datasets}}

        Unlike other research fields which have many datasets available online, such as  grasping that has over $30$ online datasets as summarised in \cite{huang2016recent},  available datasets for affordance tasks are considerably fewer. The online interactive version of this survey in footnote~\ref{url} shows a summary of the available online datasets that collect data structures to build an affordance relation for robotic tasks. In particular, in this summary we identify the task that the dataset is intended for, summarise the dataset's content and its data type, as well as provide the online location of the dataset. Given the potential to improve the agent's understanding of the task, the affordance concept has been particularly popular for object recognition, manipulation and navigation in robotic applications \cite{ardon2019learning,shu2016learning,koppula2016anticipatory}. Thus, it is not surprising that the existing online datasets focus on one of these robotic tasks. 
        
        
\section{Deployment \label{sc:deployment}}

    \begin{figure}[t!]
         \centering
              \includegraphics[width=8.9cm]{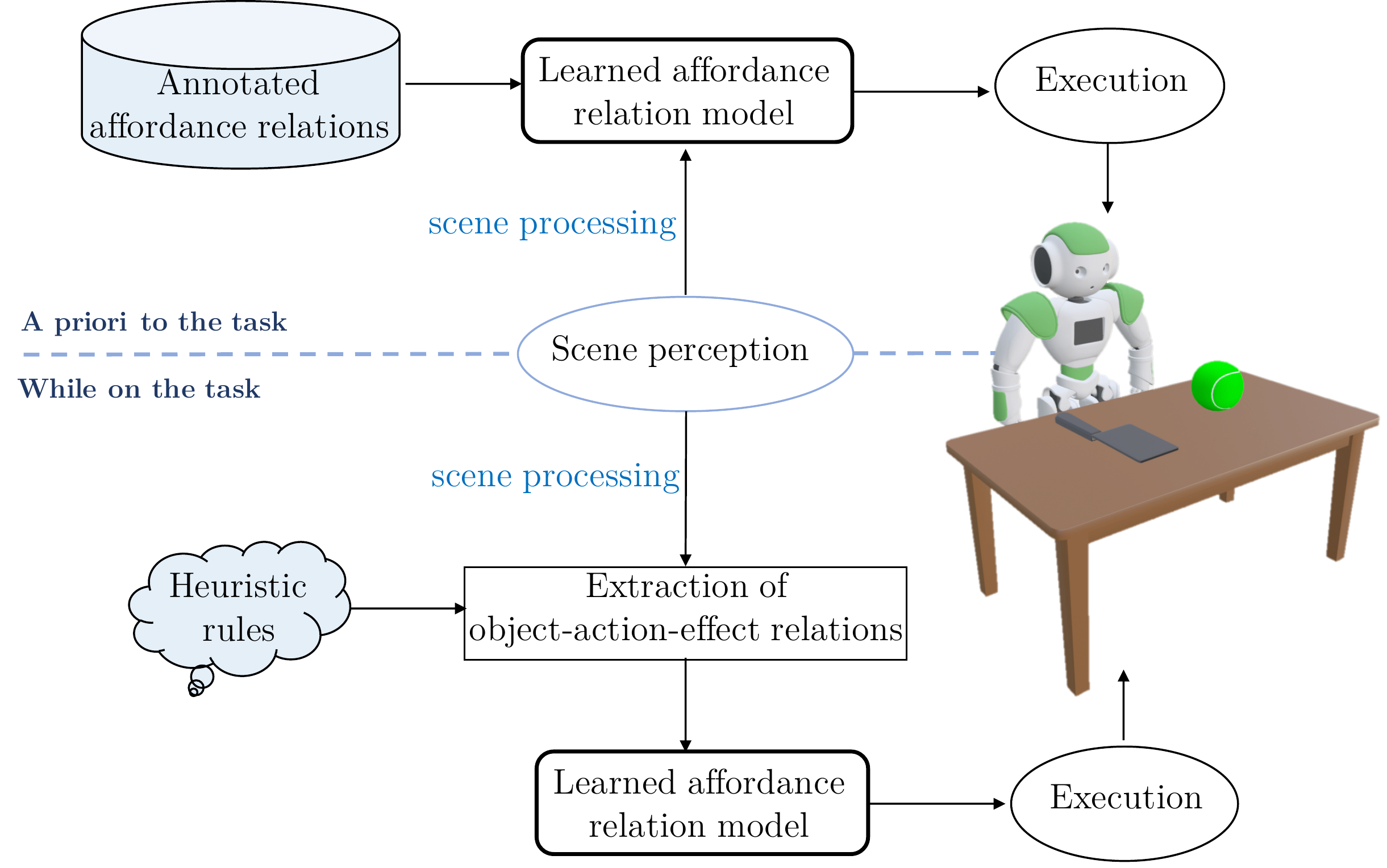}
             
             \caption{Flowcharts representing two contrasting approaches for building an affordance relation for the task of rolling an object in the scene. In both approaches, the first step is scene perception and processing (e.g., target object identification, feature extraction). In the first approach, the agent completely learns the affordance relation model prior to the task: after the data processing stage, the agent identifies a suitable affordance relation to roll the object. In the second approach, the agent has a set of rules allowing it to relate the environment perception with interactions and build an affordance relation while performing the task.}
             \label{fig:generalisation}
             \vspace{-0.3cm}
        \end{figure}
    
    In addition to the choice of data input, learning and evaluation technique reviewed in Section~\ref{sc:design_choices}, the time when data is processed and learned, with respect to the beginning of the task, affects the generalisation capabilities of the robotic agent. For example, a robot can relate pushing a ball with making it roll because it knows the ${\texttt{push} \rightarrow \texttt{roll}}$ relation before the start of the task with some probability; or a series of heuristics indicating that the features of the ball that make it round might result in the ball rolling if pushed. In the latter case, the relation ${\texttt{push} \rightarrow \texttt{roll}}$ is built with some certainty until the agent starts the affordance task, allowing the robot to adapt to previously unseen scenarios. We consider that the combination of design choices (as detailed in Section~\ref{sc:design_choices}) with the timing when the data is processed and learned determines the prior knowledge of the affordance relation. The different levels of this prior knowledge influence the system's performance in unknown environments and, as such, allow the agent to operate with different degrees of autonomy. We continue with the analogy of the robot rolling an object. In this section, we identify two general approaches to build an affordance relation in robotics and summarise them in Fig.~\ref{fig:generalisation}.
    
    
    \subsection{Building affordance relations prior to the task}
        The first general approach that we identify corresponds to methods that have full \textit{a priori task knowledge} about the possible affordance relations. Work that uses this approach usually requires less complexity in their design choices \cite{sun2014object,moldovan2014occluded,byravan2017se3,jiang2013hallucinated,koppula2016anticipatory,pieropan2014recognizing}. Namely, the information encapsulated in the affordance relation models of these methods represents their generalisation capabilities. In the literature, we identify the following three main generalisation capabilities that consider full prior knowledge of the affordance relation.
        
        \subsubsection{Using affordance relations to understand surroundings\label{sec:improve_others}}
        
            Consider the case when the robot learns a policy that indicates round objects roll. The agent is able to generalise the relation of every object in the scene that is round, by understanding object properties rather than mapping them to individual object classes (e.g., ball, apple) \cite{kim2014semantic,aksoy2015model,chu2019real,dutta2019predicting,fang2018demo2vec,luddecke2017learning}. These approaches organise their features by their `functionality'. For example, \cite{stark2008functional} generalises that features representing handles offer the possibility of grasping, as do the ones representing the surface of a bottle. For this work, learning the affordance relation of the features with actions results in superior generalisation performance for object categorisation. \cite{cruz2016training} complete a cleaning task where the simulated robot uses reinforcement learning and a predefined set of contextual affordances with few starting actions. Having this prior information enables the system to reach higher rates of success, which is the case for \cite{cruz2016training,wang2013robot}.
            
            \cite{mar2015multi,abelha2016model} enable more precise affordance predictions for tool use scenarios. In \cite{mar2015multi}, instead of learning a single model that tries to relate all the possible variables in an affordance, the robot learns a separate affordance model for each set of tools and corresponding grasping poses sharing common functionality, thus categorising tool handles and poses. Along the same lines, \cite{castellini2011using} propose the use of grasping motor data (i.e., kinematic grasping data obtained from human demonstrations) to encode the affordances of an object, and then to use this representation on similar objects to improve object recognition.
            
         \subsubsection{Affordance relation models that consider perturbations \label{sec:adapting}}
         
            Work in this sub-category learns a single model for affordance relations and keeps it fixed for the rest of the robotics task. These approaches try to achieve the task with what they know about the affordance relation in the presence of perturbations or changes in the environment. For instance, say that a robot has a prior on the ${\texttt{push} \rightarrow \texttt{roll}}$ relationship, however, when attempting to perform the task it is not able to reach the target object. As a safety policy measure, the robot knows how to find a spatula that can help it reach the target object. In this case, there are no new relations learned \cite{pandey2013affordance,fallon2015architecture,diana2013deformable,marion2017director}. As a result, the actions become rules that are queried at execution time.
            
            Some approaches in this category attempt to perform multi-step predictions based on known affordance relations, such as \cite{kroemer2011flexible,price2016affordance,cutsuridis2013cognitive,veres2020incorporating}. These methods often add a planning layer that allows them to achieve goal-oriented tasks by adapting to changes in the environment. Other work such as \cite{sun2010learning,lewis2005foot,kostavelis2012collision} and \cite{moldovan2014occluded} perform navigation and grasping application tasks, respectively, in cluttered environments. They do so in a scenario with many objects where the purpose is to identify the most suitable object for a pre-defined task. In \cite{sun2010learning,kostavelis2012collision,dogar2007primitive,saputra2019dynamic}, the goal is to arrive at a destination while choosing to push or nudge objects on the way, while in \cite{moldovan2014occluded}, the goal is to find an object that might be occluded on a shelf to achieve a queried action. \cite{wang2013robot,pandey2013affordance} use templates of interpretative triplets, containing the affordance relation components.
        
        \subsubsection{Affordance relations for multiple objects and agents \label{sec:multi}}
        
            In an environment where there is a ball and a spatula, the robot associates the ball as the object that rolls, and the spatula as the object to grasp and use to push the ball so that it rolls. In this case, the agent has a prior affordance relation model that enables it to associate multiple object affordances to achieve one task \cite{koppula2016anticipatory,pieropan2014recognizing,jiang2013hallucinated,chan2020affordance,shu2017learning,thermos2017deep}. Work that considers \textit{multi-object affordance relations} associates multiple objects in the scene \cite{ruiz2018can,chu2019learning,kaiser2016towards,varadarajan2012afrob,sun2014object,moldovan2014occluded,jiang2013hallucinated,moldovan2018relational}. They model the concept of object co-occurrence by calculating the probability of an object on a shelf being of a particular type and having a specific affordance, given that on the same shelf there are objects of a certain type. Others consider \textit{multi-agent affordance relations} in the same environment \cite{price2016affordance,song2015task}. For example, both approaches design a system where they consider the action capabilities of manipulating the objects among different agents and across places. \cite{song2015task} developed a framework by stages composed of \cite{song2013predicting,song2015task} that makes sure a robotic end-effector properly hands over an object.
    
    \subsubsection{Building affordance relations while on the task}
    
        The second general approach we found in the literature is to build the affordance relation \textit{while on the task}. This approach requires less prior knowledge and design choices that adapt to partially new environments \cite{lopes2007affordance,ugur2015staged,gonccalves2014learning,tikhanoff2013exploring,ugur2009affordance,dehban2016denoising,baleia2015exploiting,montesano2009learning}. Given that these approaches create affordance relations online, they are able to generalise to new environments. We identify two ways in which such methodologies generalise to novel scenarios.
        
        \subsubsection{Updating models with new affordance relations \label{sec:updating}}
        
            Following the example of the robot rolling an object on the scene, a possibility is that the robot knows the object rolls, from previous experience or from a tutor's demonstration. In this setup, the agent learns that poking the object from different directions affords pushing or pulling the object. As a result, the agent can extend the affordance relation model to include the new action possibilities \cite{gonccalves2014learning,tikhanoff2013exploring,ugur2009affordance,dehban2016denoising,baleia2015exploiting,montesano2009learning}. 
            
            Methods in this sub-category learn and update a model using demonstrations from a tutor or trial and error techniques. Using \textit{demonstrations}, especially for the robotics task of grasping, we find methods that exploit the benefits of \ac{LbD} to build the affordance relation model \cite{bekiroglu2013predicting,chan2014determining,ridge2013action}. For example, \cite{ridge2013action} proposes a self-supervised method that encapsulates features of the objects before and after being pushed. The features then serve as a base for the robot to know where to push new objects and create more affordance relations with other objects based on their ability to be pushed.
            
            For methods using \textit{trial and error}, they combine pre-learned affordance relation models with exploration to assess the effects of an action \cite{antunes2016human,bozcuouglu2019continuous,seker2019deep,ugur2015refining}.
        
        \subsubsection{From primitive actions to compound behaviours \label{sec:from_primitive_to_complex}}
        
            Another option available to the agent for making the object roll is to learn the combination of basic motions, such as reaching and pushing, that lead to the desired rolling outcome while performing the task,
            rather than having a prior affordance relation
           \cite{hermans2013decoupling,bonaiuto2015learning}. The approaches in this section propose a framework that allows the robot to explore and learn an affordance relation model using primitive actions as the backbone. A set of heuristic rules are then put in place to guide the robot to compose actions and associate them with a target object and effects \cite{kaiser2016towards,ugur2015staged}. These frameworks learn high-level behaviours, however, questions such as \textit{how does a robot learn to pull an object towards itself?} or \textit{how does the robot learn that spherical objects roll while a cube only slides when pushed?} concern learning of primitive actions at a control level. Some approaches learn the parameters to basic controller primitive actions to generalise to new robotic tasks by combining visual and tactile information and testing the heuristic model in a trial and error stage \cite{hermans2013decoupling,stoytchev2008learning,bonaiuto2015learning}. 
        
\section{Limitations of Affordances in Robotics \label{sc:limitations}}

  As a summary, in Section~\ref{sc:design_choices} we outlined the different design choices for data input, learning methods and evaluation of the affordance relation. In Section~\ref{sc:deployment} we described how the interaction of the design choices and the time when the affordance relation is built, with respect to the start of the task (i.e., prior knowledge of the affordance relation), influences the generalisation capabilities of the system. In particular, this correlation defines how well the methods perform on unseen environments. The diversity of design choices and prior knowledge of the affordance relation provides the field with great potential to adapt to many robotic applications. Nonetheless, this variety also makes it difficult to define standards in the field. The objective of this section is to identify and summarise the strengths (see Section~\ref{ssc:strenght}) and weaknesses (see Section~\ref{ssc:limitations}) as found in the reviewed literature.

    \subsection{Strengths \label{ssc:strenght}}
        \subsubsection{\textbf{Generalisation to novel setups}}
            The idea of identifying target objects by their functionality rather than by their categorical classes provides a system with the ability to perform the same task on similar objects. For example, an agent can learn that a bottle and a mug share features that afford pouring. Many approaches in the literature exploit the concept of generalisation to improve their object recognition and categorisation tasks \cite{aksoy2015model,kroemer2012kernel,mar2015multi,abelha2016model}.
        \subsubsection{\textbf{Potential for autonomous learning}}
           As explained in Section~\ref{sc:formalism}, different formalisms define affordances as a co-defining relation between a target object, an action to be applied on the object and an effect that evaluates such actions. As such, the concept can be conceived as a closed loop that offers the potential for the system to learn by itself with little human intervention. In the reviewed literature, we find work that has some prior on the affordance relation and that employs it to learn new relations while on the task \cite{gonccalves2014learning,ugur2009affordance,lopes2007affordance,song2010learning,ugur2015staged}.

    \subsection{Weaknesses \label{ssc:limitations}}
  
        \subsubsection{\textbf{Datasets}\label{ssc:datasets}}
            Using the concept of affordances to bias the learning of an artificial intelligent agent is relatively new. As detailed in Section~\ref{sc:formalism}, all the strategies account for target objects, actions and effects to create an affordance relation model. Nonetheless, it is difficult to find common ground for a dataset that contains these three elements and satisfies the needs of the different tasks and generalisation requirements. Unlike other self-contained research fields, such as grasping, manipulation and object recognition, affordances in robotics have few available online datasets. Most of the existing datasets are limited to (i)~a single affordance associated with an object, and (ii)~assume that this single affordance is true regardless of the context of the object. The usual approach to perform  affordance tasks is to collect a motion that represents an action. A natural step towards fast-forwarding data collection would be the design and implementation of a data collection interface. This interface would facilitate the annotation of objects with the corresponding actions to perform a robotics task. Certainly, there needs to be a consensus in the field regarding the requirements of such centralised datasets for specific robotic applications. This type of agreement would help to facilitate benchmarking different types of affordance relation models.

        \subsubsection{\textbf{Metrics}\label{ssc:metrics}}
       
           It is fundamental for the progress of the affordance concept as a learning bias to standardise metrics that reflect the performance of an affordance-aware agent in different tasks. A summary of the diversity of metrics used in the field is detailed in Section~\ref{ssc:metrics}. Given that most of the contributions lay in collaborative tasks with other agents and improving generalisation performance, an interesting approach would be to measure the similarity of the actions taken by the system with those a human would execute. For example, it could be interesting to measure the differences in the trajectories executed to achieve a task. Such differences could be measured in terms of distances in the point distributions of the trajectories, or entropy of the trajectories as a whole. Options such as the Hausdorff distance and the Kullback-Leibler divergence are interesting to explore. The Hausdorff distance measures how similar or close two sets of points are, and the Kullback-Leibler divergence measures how one probability distribution is different from a second one. Including such evaluations could be a good assessment of performance on a robotic task in relation to ground truth data, regardless of learning.



\section{Conclusions and Future Directions \label{sc:openquestions}}

    In this survey, we explored the literature for approaches that included affordances in the execution of AI and robotic tasks, and identified common ground for building affordance relations. In contrast to previous reviews of affordances, we provide guidance on design decisions and how the concept can be used to guide policy learning to boost the agent's performance. First, we  summarised affordance formalisms in Section~\ref{sc:formalism}, where we found that affordance relations are built using three common elements: target object, action and effect. Then, we outlined the design choices in Section~\ref{sc:design_choices} and the possible generalisation schemes in Section~\ref{sc:deployment}. In Section~\ref{sc:limitations}, we discussed several problems in the field. Given the relatively new usage of affordances to boost the agent's generalisation capabilities, there are interesting opportunities for future improvements. Next, we outline possible areas where research contributions can be made, based on our survey of the reviewed literature.

    \subsection{Design choices and their influence \label{ssc:summary}}
\begin{figure}[t!]
            \centering
              \includegraphics[width=8.9cm]{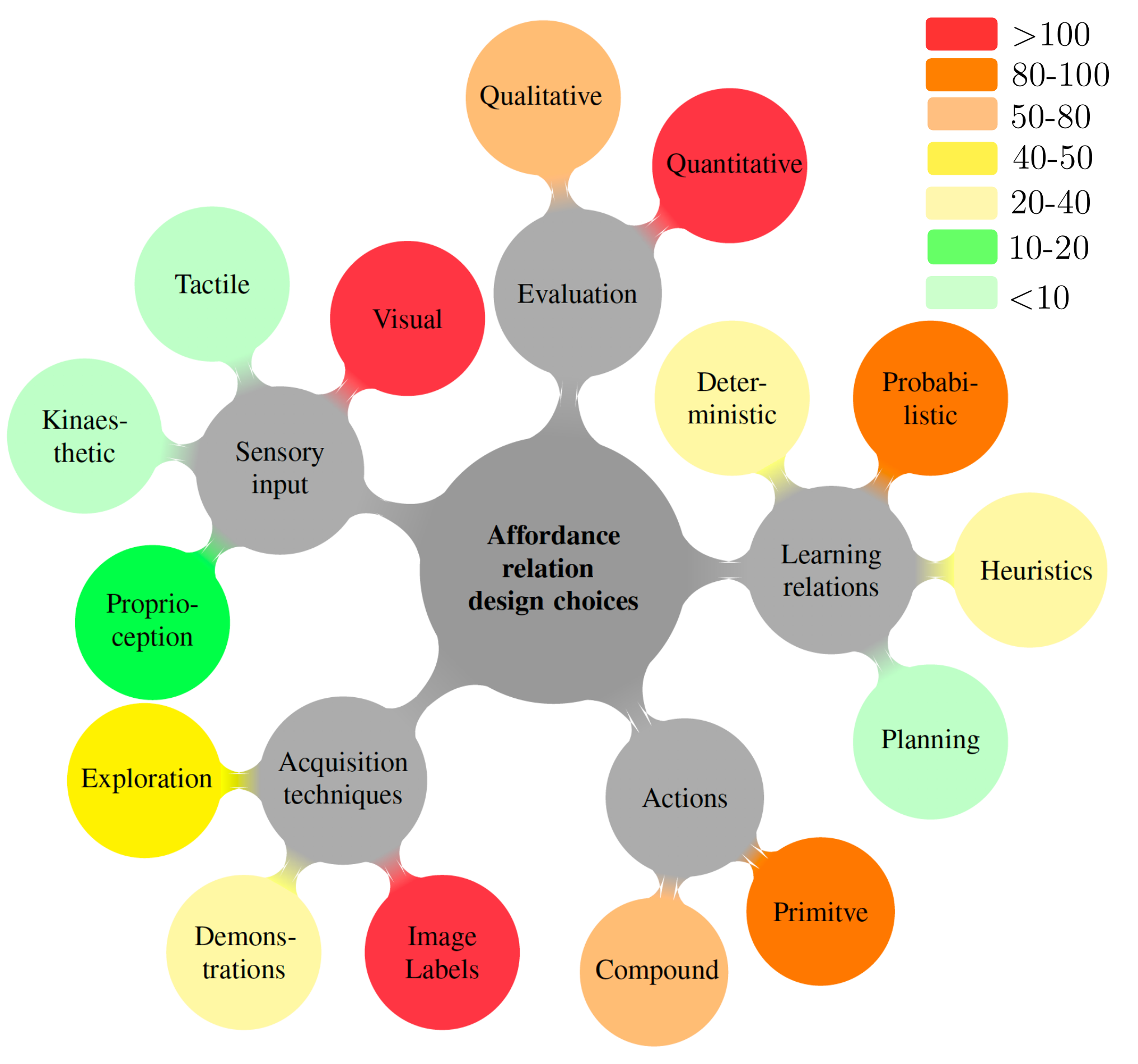}
           
            \caption{Population map of the design choices for work in the reviewed literature that includes affordances in  robotic tasks. The warmer the colour (red) the more the work uses that element to design their affordance relation model, while the colder the colour (green), the less that element is used.}
            \label{fig:coverage}
            \vspace{-0.3cm}
        \end{figure}

        As previously mentioned, the choice of data input and learning time influence the performance of the methods, often in different ways. Nonetheless, some design choices have been explored more than others, thus leaving room for research into the generalisation capabilities that can be achieved. Fig.~\ref{fig:coverage} shows a coverage map of the reviewed literature spread over the different design choices. The warmer the colour, the more the strategy is used across the literature. For example, most of the work emphasises the learning of primitive actions as affordances (i.e., push, grasp, lift, among others), using visual perception and image labels to identify an affordance per target object, building the affordance relation probabilistically. 
        On the opposite side, colder coloured elements indicate that there are very few approaches that exploit learning affordance trajectories in the form of motions (using kinaesthetic sensing), as well as those that exploit a multi-step prediction to achieve the tasks in a planning manner. Certainly, some of these components are highly dependent on hardware robustness more than others. Nonetheless, studying such aspects in greater depth would improve their inclusion in robotics tasks as well as provide valuable insights for collaboration activities and task replication across different agents.
    
    \subsection{Data acquisition and processing correlation \label{sec:challenges_is}}
        Given that the field requires different types of data (i.e., target object, action and effects), there is an inherent need to cross-correlate data structures. These associations can be further enriched by diversifying data acquisition and processing. For example, knowing that a cup in a kitchen is likely to afford pouring liquid while in a bathroom might also serve as a toothbrush holder requires the agent to be able to relate not only object features but also features related to an agent's surroundings. Moreover, the data acquisition task should not be limited to visual object features alone but should also account for the object's material, texture and other physical properties to enhance the agent's interaction with the object. By using such cross-correlation, affordance relation models could provide the agents with the ability to generalise affordances as associations of objects' visual and physical characteristics, as well as with the surrounding context.
     
    \subsection{Autonomous behaviour learning \label{sec:challenges_ma}}
        At present, the idea of including the concept of affordances in AI tasks has centred on performing one task at the time. For instance, many approaches detect one object affordance, such as a glass affords pouring, but rarely proceed with actually performing the pouring task. The field would benefit from a methodology that is able to unify associations of target objects with a library of actions and an online evaluation of the effects. This association would allow an agent to obtain feedback on the performance of the task and rectify it online. Such a method would open doors for autonomously concatenating goal-oriented tasks (e.g., preparing a recipe, cleaning dishes, etc.), and for exploring the feasibility of subsequent motion controllers and the interpretability of natural language instructions in large-scale tasks.

{
\fontsize{8}{8}\selectfont
\bibliographystyle{named}
\bibliography{ijcai21}

\begin{thebibliography}{}

\bibitem[\protect\citeauthoryear{Abelha \bgroup \em et al.\egroup
  }{2016}]{abelha2016model}
Paulo Abelha, Frank Guerin, and Markus Schoeler.
\newblock A model-based approach to finding substitute tools in {3D} vision
  data.
\newblock In {\em IEEE International Conference on Robotics and Automation},
  pages 2471--2478. IEEE, 2016.

\bibitem[\protect\citeauthoryear{Aksoy \bgroup \em et al.\egroup
  }{2015}]{aksoy2015model}
Eren~Erdal Aksoy, Minija Tamosiunaite, and Florentin W{\"o}rg{\"o}tter.
\newblock Model-free incremental learning of the semantics of manipulation
  actions.
\newblock {\em Robotics and Autonomous Systems}, 71:118--133, 2015.

\bibitem[\protect\citeauthoryear{Aldoma \bgroup \em et al.\egroup
  }{2012}]{aldoma2012supervised}
Aitor Aldoma, Federico Tombari, and Markus Vincze.
\newblock Supervised learning of hidden and non-hidden 0-order affordances and
  detection in real scenes.
\newblock In {\em IEEE International Conference on Robotics and Automation},
  pages 1732--1739. IEEE, 2012.

\bibitem[\protect\citeauthoryear{Antunes \bgroup \em et al.\egroup
  }{2016}]{antunes2016human}
Alexandre Antunes, Lorenzo Jamone, Giovanni Saponaro, Alexandre Bernardino, and
  Rodrigo Ventura.
\newblock From human instructions to robot actions: Formulation of goals,
  affordances and probabilistic planning.
\newblock In {\em IEEE International Conference on Robotics and Automation},
  pages 5449--5454. IEEE, 2016.

\bibitem[\protect\citeauthoryear{Ard{\'o}n \bgroup \em et al.\egroup
  }{2019}]{ardon2019learning}
Paola Ard{\'o}n, {\`E}ric Pairet, Ronald~PA Petrick, Subramanian Ramamoorthy,
  and Katrin~S Lohan.
\newblock Learning grasp affordance reasoning through semantic relations.
\newblock {\em IEEE Robotics and Automation Letters}, 4(4):4571--4578, 2019.

\bibitem[\protect\citeauthoryear{Ard{\'o}n \bgroup \em et al.\egroup
  }{2020}]{ardon2020self}
Paola Ard{\'o}n, {\`E}ric Pairet, Yvan Petillot, Ronald~PA Petrick, Subramanian
  Ramamoorthy, and Katrin~S Lohan.
\newblock Self-assessment of grasp affordance transfer.
\newblock In {\em IEEE/RSJ International Conference on Intelligent Robots and
  Systems}. IEEE, 2020.

\bibitem[\protect\citeauthoryear{Asada \bgroup \em et al.\egroup
  }{2009}]{asada2009cognitive}
Minoru Asada, Koh Hosoda, Yasuo Kuniyoshi, Hiroshi Ishiguro, Toshio Inui,
  Yuichiro Yoshikawa, Masaki Ogino, and Chisato Yoshida.
\newblock Cognitive developmental robotics: A survey.
\newblock {\em IEEE Transactions on Autonomous Mental Development},
  1(1):12--34, 2009.

\bibitem[\protect\citeauthoryear{Baleia \bgroup \em et al.\egroup
  }{2015}]{baleia2015exploiting}
Jos{\'e} Baleia, Pedro Santana, and Jos{\'e} Barata.
\newblock On exploiting haptic cues for self-supervised learning of depth-based
  robot navigation affordances.
\newblock {\em Journal of Intelligent \& Robotic Systems}, 80(3-4):455--474,
  2015.

\bibitem[\protect\citeauthoryear{Barck-Holst \bgroup \em et al.\egroup
  }{2009}]{barck2009learning}
Carl Barck-Holst, Maria Ralph, Fredrik Holmar, and Danica Kragic.
\newblock Learning grasping affordance using probabilistic and ontological
  approaches.
\newblock In {\em International Conference on Advanced Robotics}, pages 1--6.
  IEEE, 2009.

\bibitem[\protect\citeauthoryear{Bekiroglu \bgroup \em et al.\egroup
  }{2013}]{bekiroglu2013predicting}
Yasemin Bekiroglu, Christian Smith, Yiannis Karayiannidis, Danica Kragic,
  et~al.
\newblock Predicting slippage and learning manipulation affordances through
  gaussian process regression.
\newblock In {\em IEEE-RAS International Conference on Humanoid Robots}, pages
  462--468. IEEE, 2013.

\bibitem[\protect\citeauthoryear{Bierbaum \bgroup \em et al.\egroup
  }{2009}]{bierbaum2009grasp}
Alexander Bierbaum, Matthias Rambow, Tamim Asfour, and R{\"u}diger Dillmann.
\newblock Grasp affordances from multi-fingered tactile exploration using
  dynamic potential fields.
\newblock In {\em IEEE-RAS International Conference on Humanoid Robots}, pages
  168--174. IEEE, 2009.

\bibitem[\protect\citeauthoryear{Bohg \bgroup \em et al.\egroup
  }{2017}]{bohg2017interactive}
Jeannette Bohg, Karol Hausman, Bharath Sankaran, Oliver Brock, Danica Kragic,
  Stefan Schaal, and Gaurav~S Sukhatme.
\newblock Interactive perception: Leveraging action in perception and
  perception in action.
\newblock {\em IEEE Transactions on Robotics}, 33(6):1273--1291, 2017.

\bibitem[\protect\citeauthoryear{Bonaiuto and
  Arbib}{2015}]{bonaiuto2015learning}
James Bonaiuto and Michael~A Arbib.
\newblock Learning to grasp and extract affordances: the integrated learning of
  grasps and affordances ({ILGA}) model.
\newblock {\em Biological cybernetics}, 109(6):639--669, 2015.

\bibitem[\protect\citeauthoryear{Bozcuo{\u{g}}lu \bgroup \em et al.\egroup
  }{2019}]{bozcuouglu2019continuous}
Asil~Kaan Bozcuo{\u{g}}lu, Yuki Furuta, Kei Okada, Michael Beetz, and Masayuki
  Inaba.
\newblock Continuous modeling of affordances in a symbolic knowledge base.
\newblock In {\em IEEE/RSJ International Conference on Intelligent Robots and
  Systems}, pages 5452--5458. IEEE, 2019.

\bibitem[\protect\citeauthoryear{Byravan and Fox}{2017}]{byravan2017se3}
Arunkumar Byravan and Dieter Fox.
\newblock Se3-nets: Learning rigid body motion using deep neural networks.
\newblock In {\em IEEE International Conference on Robotics and Automation},
  pages 173--180. IEEE, 2017.

\bibitem[\protect\citeauthoryear{Cai \bgroup \em et al.\egroup
  }{2019}]{cai2019metagrasp}
Junhao Cai, Hui Cheng, Zhanpeng Zhang, and Jingcheng Su.
\newblock Metagrasp: Data efficient grasping by affordance interpreter network.
\newblock In {\em International Conference on Robotics and Automation}, pages
  4960--4966. IEEE, 2019.

\bibitem[\protect\citeauthoryear{Castellini \bgroup \em et al.\egroup
  }{2011}]{castellini2011using}
Claudio Castellini, Tatiana Tommasi, Nicoletta Noceti, Francesca Odone, and
  Barbara Caputo.
\newblock Using object affordances to improve object recognition.
\newblock {\em IEEE Transactions on Autonomous Mental Development},
  3(3):207--215, 2011.

\bibitem[\protect\citeauthoryear{Chan \bgroup \em et al.\egroup
  }{2014}]{chan2014determining}
Wesley~P Chan, Yohei Kakiuchi, Kei Okada, and Masayuki Inaba.
\newblock Determining proper grasp configurations for handovers through
  observation of object movement patterns and inter object interactions during
  usage.
\newblock In {\em IEEE/RSJ International Conference on Intelligent Robots and
  Systems}, pages 1355--1360. IEEE, 2014.

\bibitem[\protect\citeauthoryear{Chan \bgroup \em et al.\egroup
  }{2020}]{chan2020affordance}
Wesley~P Chan, Matthew~KXJ Pan, Elizabeth~A Croft, and Masayuki Inaba.
\newblock An affordance and distance minimization based method for computing
  object orientations for robot human handovers.
\newblock {\em International Journal of Social Robotics}, 12(1):143--162, 2020.

\bibitem[\protect\citeauthoryear{Chemero and
  Turvey}{2007}]{chemero2007gibsonian}
Anthony Chemero and Michael~T Turvey.
\newblock Gibsonian affordances for roboticists.
\newblock {\em Adaptive Behavior}, 15(4):473--480, 2007.

\bibitem[\protect\citeauthoryear{Chu \bgroup \em et al.\egroup
  }{2016}]{chu2016learning}
Vivian Chu, Tesca Fitzgerald, and Andrea~L Thomaz.
\newblock Learning object affordances by leveraging the combination of
  human-guidance and self-exploration.
\newblock In {\em ACM/IEEE International Conference on Human-Robot
  Interaction}, pages 221--228. IEEE, 2016.

\bibitem[\protect\citeauthoryear{Chu \bgroup \em et al.\egroup
  }{2019a}]{chu2019learning}
Fu-Jen Chu, Ruinian Xu, and Patricio~A Vela.
\newblock Learning affordance segmentation for real-world robotic manipulation
  via synthetic images.
\newblock {\em IEEE Robotics and Automation Letters}, 4(2):1140--1147, 2019.

\bibitem[\protect\citeauthoryear{Chu \bgroup \em et al.\egroup
  }{2019b}]{chu2019real}
Vivian Chu, Reymundo~A Gutierrez, Sonia Chernova, and Andrea~L Thomaz.
\newblock Real-time multisensory affordance-based control for adaptive object
  manipulation.
\newblock In {\em International Conference on Robotics and Automation}, pages
  7783--7790. IEEE, 2019.

\bibitem[\protect\citeauthoryear{Cruz \bgroup \em et al.\egroup
  }{2016}]{cruz2016training}
Francisco Cruz, Sven Magg, Cornelius Weber, and Stefan Wermter.
\newblock Training agents with interactive reinforcement learning and
  contextual affordances.
\newblock {\em IEEE Transactions on Cognitive and Developmental Systems},
  8(4):271--284, 2016.

\bibitem[\protect\citeauthoryear{Cutsuridis and
  Taylor}{2013}]{cutsuridis2013cognitive}
Vassilis Cutsuridis and John~G Taylor.
\newblock A cognitive control architecture for the perception--action cycle in
  robots and agents.
\newblock {\em Cognitive Computation}, 5(3):383--395, 2013.

\bibitem[\protect\citeauthoryear{Dehban \bgroup \em et al.\egroup
  }{2016}]{dehban2016denoising}
Atabak Dehban, Lorenzo Jamone, Adam~R Kampff, and Jos{\'e} Santos-Victor.
\newblock Denoising auto-encoders for learning of objects and tools affordances
  in continuous space.
\newblock In {\em IEEE International Conference on Robotics and Automation},
  pages 4866--4871. IEEE, 2016.

\bibitem[\protect\citeauthoryear{Diana \bgroup \em et al.\egroup
  }{2013}]{diana2013deformable}
Matteo Diana, Jean-Pierre de~la Croix, and Magnus Egerstedt.
\newblock Deformable-medium affordances for interacting with multi-robot
  systems.
\newblock In {\em IEEE/RSJ International Conference on Intelligent Robots and
  Systems}, pages 5252--5257. IEEE, 2013.

\bibitem[\protect\citeauthoryear{Do \bgroup \em et al.\egroup
  }{2018}]{do2018affordancenet}
Thanh-Toan Do, Anh Nguyen, and Ian Reid.
\newblock Affordancenet: An end-to-end deep learning approach for object
  affordance detection.
\newblock In {\em IEEE international conference on robotics and automation},
  pages 5882--5889. IEEE, 2018.

\bibitem[\protect\citeauthoryear{Dogar \bgroup \em et al.\egroup
  }{2007}]{dogar2007primitive}
Mehmet~R Dogar, Maya Cakmak, Emre Ugur, and Erol Sahin.
\newblock From primitive behaviors to goal-directed behavior using affordances.
\newblock In {\em IEEE/RSJ International Conference on Intelligent Robots and
  Systems}, pages 729--734. IEEE, 2007.

\bibitem[\protect\citeauthoryear{Dutta and
  Zielinska}{2019}]{dutta2019predicting}
Vibekananda Dutta and Teresa Zielinska.
\newblock Predicting human actions taking into account object affordances.
\newblock {\em Journal of Intelligent \& Robotic Systems}, 93(3-4):745--761,
  2019.

\bibitem[\protect\citeauthoryear{ECCV}{2014}]{workshop2}
ECCV.
\newblock Second workshop on affordances: Visual perception of affordances and
  functional visual primitives for scene analysis, 2014.

\bibitem[\protect\citeauthoryear{Fallon \bgroup \em et al.\egroup
  }{2015}]{fallon2015architecture}
Maurice Fallon, Scott Kuindersma, Sisir Karumanchi, Matthew Antone, Toby
  Schneider, Hongkai Dai, Claudia~P{\'e}rez D'Arpino, Robin Deits, Matt
  DiCicco, Dehann Fourie, et~al.
\newblock An architecture for online affordance-based perception and whole-body
  planning.
\newblock {\em Journal of Field Robotics}, 32(2):229--254, 2015.

\bibitem[\protect\citeauthoryear{Fang \bgroup \em et al.\egroup
  }{2018}]{fang2018demo2vec}
Kuan Fang, Te-Lin Wu, Daniel Yang, Silvio Savarese, and Joseph~J Lim.
\newblock Demo2{V}ec: Reasoning object affordances from online videos.
\newblock In {\em IEEE Conference on Computer Vision and Pattern Recognition},
  pages 2139--2147, 2018.

\bibitem[\protect\citeauthoryear{Gibson and
  Carmichael}{1966}]{gibson1966senses}
James~Jerome Gibson and Leonard Carmichael.
\newblock {\em The senses considered as perceptual systems}, volume~2.
\newblock Houghton Mifflin Boston, 1966.

\bibitem[\protect\citeauthoryear{Gon{\c{c}}alves \bgroup \em et al.\egroup
  }{2014}]{gonccalves2014learning}
Afonso Gon{\c{c}}alves, Jo{\~a}o Abrantes, Giovanni Saponaro, Lorenzo Jamone,
  and Alexandre Bernardino.
\newblock Learning intermediate object affordances: Towards the development of
  a tool concept.
\newblock In {\em International Conference on Development and Learning and on
  Epigenetic Robotics}, pages 482--488. IEEE, 2014.

\bibitem[\protect\citeauthoryear{Hammond}{2010}]{hammond2010affordance}
Michael Hammond.
\newblock What is an affordance and can it help us understand the use of {ICT}
  in education?
\newblock {\em Education and Information Technologies}, 15(3):205--217, 2010.

\bibitem[\protect\citeauthoryear{Hermans \bgroup \em et al.\egroup
  }{2013}]{hermans2013decoupling}
Tucker Hermans, James~M Rehg, and Aaron~F Bobick.
\newblock Decoupling behavior, perception, and control for autonomous learning
  of affordances.
\newblock In {\em IEEE International Conference on Robotics and Automation},
  pages 4989--4996. IEEE, 2013.

\bibitem[\protect\citeauthoryear{Horton \bgroup \em et al.\egroup
  }{2012}]{horton2012affordances}
Thomas~E Horton, Arpan Chakraborty, and Robert~St Amant.
\newblock Affordances for robots: a brief survey.
\newblock {\em AVANT. Pismo Awangardy Filozoficzno-Naukowej}, 2:70--84, 2012.

\bibitem[\protect\citeauthoryear{Huang \bgroup \em et al.\egroup
  }{2016}]{huang2016recent}
Yongqiang Huang, Matteo Bianchi, Minas Liarokapis, and Yu~Sun.
\newblock Recent data sets on object manipulation: A survey.
\newblock {\em Big data}, 4(4):197--216, 2016.

\bibitem[\protect\citeauthoryear{ICRA}{2019}]{workshop4}
ICRA.
\newblock 2nd edition of international workshop on computational models of
  affordance in robotic, 2019.

\bibitem[\protect\citeauthoryear{Jamone \bgroup \em et al.\egroup
  }{2016}]{jamone2016affordances}
Lorenzo Jamone, Emre Ugur, Angelo Cangelosi, Luciano Fadiga, Alexandre
  Bernardino, Justus Piater, and Jos{\'e} Santos-Victor.
\newblock Affordances in psychology, neuroscience, and robotics: A survey.
\newblock {\em IEEE Transactions on Cognitive and Developmental Systems},
  10(1):4--25, 2016.

\bibitem[\protect\citeauthoryear{Jiang \bgroup \em et al.\egroup
  }{2013}]{jiang2013hallucinated}
Yun Jiang, Hema Koppula, and Ashutosh Saxena.
\newblock Hallucinated humans as the hidden context for labeling 3d scenes.
\newblock In {\em IEEE Conference on Computer Vision and Pattern Recognition},
  pages 2993--3000, 2013.

\bibitem[\protect\citeauthoryear{Kaiser \bgroup \em et al.\egroup
  }{2015}]{kaiser2015validation}
Peter Kaiser, Markus Grotz, Eren~E Aksoy, Martin Do, Nikolaus Vahrenkamp, and
  Tamim Asfour.
\newblock Validation of whole-body loco-manipulation affordances for
  pushability and liftability.
\newblock In {\em IEEE-RAS International Conference on Humanoid Robots}, pages
  920--927. IEEE, 2015.

\bibitem[\protect\citeauthoryear{Kaiser \bgroup \em et al.\egroup
  }{2016}]{kaiser2016towards}
Peter Kaiser, Eren~E Aksoy, Markus Grotz, and Tamim Asfour.
\newblock Towards a hierarchy of loco-manipulation affordances.
\newblock In {\em IEEE/RSJ International Conference on Intelligent Robots and
  Systems}, pages 2839--2846. IEEE, 2016.

\bibitem[\protect\citeauthoryear{Katz \bgroup \em et al.\egroup
  }{2014}]{katz2014perceiving}
Dov Katz, Arun Venkatraman, Moslem Kazemi, J~Andrew Bagnell, and Anthony
  Stentz.
\newblock Perceiving, learning, and exploiting object affordances for
  autonomous pile manipulation.
\newblock {\em Autonomous Robots}, 37(4):369--382, 2014.

\bibitem[\protect\citeauthoryear{Kim and Sukhatme}{2014}]{kim2014semantic}
David~Inkyu Kim and Gaurav~S Sukhatme.
\newblock Semantic labeling of {3D} point clouds with object affordance for
  robot manipulation.
\newblock In {\em IEEE International Conference on Robotics and Automation},
  pages 5578--5584, 2014.

\bibitem[\protect\citeauthoryear{Koppula and
  Saxena}{2014}]{koppula2014physically}
Hema~S Koppula and Ashutosh Saxena.
\newblock Physically grounded spatio-temporal object affordances.
\newblock In {\em European Conference on Computer Vision}, pages 831--847.
  Springer, 2014.

\bibitem[\protect\citeauthoryear{Koppula \bgroup \em et al.\egroup
  }{2016}]{koppula2016anticipatory}
Hema~S Koppula, Ashesh Jain, and Ashutosh Saxena.
\newblock Anticipatory planning for human-robot teams.
\newblock In {\em Experimental Robotics}, pages 453--470. Springer, 2016.

\bibitem[\protect\citeauthoryear{Kostavelis \bgroup \em et al.\egroup
  }{2012}]{kostavelis2012collision}
Ioannis Kostavelis, Lazaros Nalpantidis, and Antonios Gasteratos.
\newblock Collision risk assessment for autonomous robots by offline
  traversability learning.
\newblock {\em Robotics and Autonomous Systems}, 60(11):1367--1376, 2012.

\bibitem[\protect\citeauthoryear{Kroemer and
  Peters}{2011}]{kroemer2011flexible}
Oliver Kroemer and Jan Peters.
\newblock A flexible hybrid framework for modeling complex manipulation tasks.
\newblock In {\em IEEE International Conference on Robotics and Automation},
  pages 1856--1861. IEEE, 2011.

\bibitem[\protect\citeauthoryear{Kroemer \bgroup \em et al.\egroup
  }{2012}]{kroemer2012kernel}
Oliver Kroemer, Emre Ugur, Erhan Oztop, and Jan Peters.
\newblock A kernel-based approach to direct action perception.
\newblock In {\em IEEE International Conference on Robotics and Automation},
  pages 2605--2610. IEEE, 2012.

\bibitem[\protect\citeauthoryear{Kr{\"u}ger \bgroup \em et al.\egroup
  }{2011}]{kruger2011object}
Norbert Kr{\"u}ger, Christopher Geib, Justus Piater, Ronald Petrick, Mark
  Steedman, Florentin W{\"o}rg{\"o}tter, Ale{\v{s}} Ude, Tamim Asfour, Dirk
  Kraft, Damir Omr{\v{c}}en, et~al.
\newblock Object--action complexes: Grounded abstractions of sensory--motor
  processes.
\newblock {\em Robotics and Autonomous Systems}, 59(10):740--757, 2011.

\bibitem[\protect\citeauthoryear{Lewis \bgroup \em et al.\egroup
  }{2005}]{lewis2005foot}
M~Anthony Lewis, Hyo-Kyung Lee, and Aftab Patla.
\newblock Foot placement selection using non-geometric visual properties.
\newblock {\em The International Journal of Robotics Research}, 24(7):553--561,
  2005.

\bibitem[\protect\citeauthoryear{Liu \bgroup \em et al.\egroup
  }{2019}]{liu2019learning}
Chunfang Liu, Bin Fang, Fuchun Sun, Xiaoli Li, and Wenbing Huang.
\newblock Learning to grasp familiar objects based on experience and objects’
  shape affordance.
\newblock {\em IEEE Transactions on Systems, Man, and Cybernetics: Systems},
  49(12):2710--2723, 2019.

\bibitem[\protect\citeauthoryear{Lopes \bgroup \em et al.\egroup
  }{2007}]{lopes2007affordance}
Manuel Lopes, Francisco~S Melo, and Luis Montesano.
\newblock Affordance-based imitation learning in robots.
\newblock In {\em IEEE/RSJ International Conference on Intelligent Robots and
  Systems}, pages 1015--1021. IEEE, 2007.

\bibitem[\protect\citeauthoryear{Luddecke and
  Worgotter}{2017}]{luddecke2017learning}
Timo Luddecke and Florentin Worgotter.
\newblock Learning to segment affordances.
\newblock In {\em IEEE International Conference on Computer Vision Workshops},
  pages 769--776, 2017.

\bibitem[\protect\citeauthoryear{Mar \bgroup \em et al.\egroup
  }{2015}]{mar2015multi}
Tanis Mar, Vadim Tikhanoff, Giorgio Metta, and Lorenzo Natale.
\newblock Multi-model approach based on {3D} functional features for tool
  affordance learning in robotics.
\newblock In {\em IEEE-RAS International Conference on Humanoid Robots}, pages
  482--489. IEEE, 2015.

\bibitem[\protect\citeauthoryear{Marion \bgroup \em et al.\egroup
  }{2017}]{marion2017director}
Pat Marion, Maurice Fallon, Robin Deits, Andr{\'e}s Valenzuela, Claudia
  P{\'e}rez~D'Arpino, Greg Izatt, Lucas Manuelli, Matt Antone, Hongkai Dai,
  Twan Koolen, et~al.
\newblock Director: A user interface designed for robot operation with shared
  autonomy.
\newblock {\em Journal of Field Robotics}, 34(2):262--280, 2017.

\bibitem[\protect\citeauthoryear{McGrenere and
  Ho}{2000}]{mcgrenere2000affordances}
Joanna McGrenere and Wayne Ho.
\newblock Affordances: Clarifying and evolving a concept.
\newblock In {\em Graphics interface}, volume 2000, pages 179--186, 2000.

\bibitem[\protect\citeauthoryear{Min \bgroup \em et al.\egroup
  }{2016}]{min2016affordance}
Huaqing Min, Chang’an Yi, Ronghua Luo, Jinhui Zhu, and Sheng Bi.
\newblock Affordance research in developmental robotics: a survey.
\newblock {\em IEEE Transactions on Cognitive and Developmental Systems},
  8(4):237--255, 2016.

\bibitem[\protect\citeauthoryear{Moldovan and
  De~Raedt}{2014}]{moldovan2014occluded}
Bogdan Moldovan and Luc De~Raedt.
\newblock Occluded object search by relational affordances.
\newblock In {\em IEEE International Conference on Robotics and Automation},
  pages 169--174. IEEE, 2014.

\bibitem[\protect\citeauthoryear{Moldovan \bgroup \em et al.\egroup
  }{2012}]{moldovan2012learning}
Bogdan Moldovan, Plinio Moreno, Martijn van Otterlo, Jos{\'e} Santos-Victor,
  and Luc De~Raedt.
\newblock Learning relational affordance models for robots in multi-object
  manipulation tasks.
\newblock In {\em IEEE International Conference on Robotics and Automation},
  pages 4373--4378. IEEE, 2012.

\bibitem[\protect\citeauthoryear{Moldovan \bgroup \em et al.\egroup
  }{2018}]{moldovan2018relational}
Bogdan Moldovan, Plinio Moreno, Davide Nitti, Jos{\'e} Santos-Victor, and Luc
  De~Raedt.
\newblock Relational affordances for multiple-object manipulation.
\newblock {\em Autonomous Robots}, 42(1):19--44, 2018.

\bibitem[\protect\citeauthoryear{Montesano and
  Lopes}{2009}]{montesano2009learning}
Luis Montesano and Manuel Lopes.
\newblock Learning grasping affordances from local visual descriptors.
\newblock In {\em IEEE International Conference on Development and Learning},
  pages 1--6. IEEE, 2009.

\bibitem[\protect\citeauthoryear{Montesano \bgroup \em et al.\egroup
  }{2007}]{montesano2007modeling}
Luis Montesano, Manuel Lopes, Alexandre Bernardino, and Jose Santos-Victor.
\newblock Modeling affordances using bayesian networks.
\newblock In {\em IEEE/RSJ International Conference on Intelligent Robots and
  Systems}, pages 4102--4107. IEEE, 2007.

\bibitem[\protect\citeauthoryear{Myers \bgroup \em et al.\egroup
  }{2015}]{myers2015affordance}
Austin Myers, Ching~L Teo, Cornelia Ferm{\"u}ller, and Yiannis Aloimonos.
\newblock Affordance detection of tool parts from geometric features.
\newblock In {\em IEEE International Conference on Robotics and Automation},
  pages 1374--1381. IEEE, 2015.

\bibitem[\protect\citeauthoryear{Nguyen \bgroup \em et al.\egroup
  }{2017}]{nguyen2017object}
Anh Nguyen, Dimitrios Kanoulas, Darwin~G Caldwell, and Nikos~G Tsagarakis.
\newblock Object-based affordances detection with convolutional neural networks
  and dense conditional random fields.
\newblock In {\em IEEE/RSJ International Conference on Intelligent Robots and
  Systems}, pages 5908--5915. IEEE, 2017.

\bibitem[\protect\citeauthoryear{Norman}{1988}]{norman1988psychology}
Donald~A Norman.
\newblock The psychology of everyday things, 1988.

\bibitem[\protect\citeauthoryear{Pandey and Alami}{2013}]{pandey2013affordance}
Amit~Kumar Pandey and Rachid Alami.
\newblock Affordance graph: A framework to encode perspective taking and effort
  based affordances for day-to-day human-robot interaction.
\newblock In {\em IEEE/RSJ International Conference on Intelligent Robots and
  Systems}, pages 2180--2187. IEEE, 2013.

\bibitem[\protect\citeauthoryear{Pieropan \bgroup \em et al.\egroup
  }{2014}]{pieropan2014recognizing}
Alessandro Pieropan, Carl~Henrik Ek, and Hedvig Kjellstr{\"o}m.
\newblock Recognizing object affordances in terms of spatio-temporal
  object-object relationships.
\newblock In {\em International Conference on Humanoid Robots}, pages 52--58.
  IEEE, 2014.

\bibitem[\protect\citeauthoryear{Price \bgroup \em et al.\egroup
  }{2016}]{price2016affordance}
Andrew Price, Stephen Balakirsky, Aaron Bobick, and Henrik Christensen.
\newblock Affordance-feasible planning with manipulator wrench spaces.
\newblock In {\em IEEE International Conference on Robotics and Automation},
  pages 3979--3986. IEEE, 2016.

\bibitem[\protect\citeauthoryear{Ridge and Ude}{2013}]{ridge2013action}
Barry Ridge and Ales Ude.
\newblock Action-grounded push affordance bootstrapping of unknown objects.
\newblock In {\em IEEE/RSJ International Conference on Intelligent Robots and
  Systems}, pages 2791--2798. IEEE, 2013.

\bibitem[\protect\citeauthoryear{RSS}{2014}]{workshop1}
RSS.
\newblock First workshop on affordances: Affordances in vision for cognitive
  robotics, 2014.

\bibitem[\protect\citeauthoryear{RSS}{2018}]{workshop3}
RSS.
\newblock International workshop on computational models ofaffordance in
  robotic, 2018.

\bibitem[\protect\citeauthoryear{Ruiz and Mayol-Cuevas}{2018}]{ruiz2018can}
Eduardo Ruiz and Walterio Mayol-Cuevas.
\newblock Where can {I} do this? {G}eometric affordances from a single example
  with the interaction tensor.
\newblock In {\em IEEE International Conference on Robotics and Automation},
  pages 2192--2199. IEEE, 2018.

\bibitem[\protect\citeauthoryear{{\c{S}}ahin \bgroup \em et al.\egroup
  }{2007}]{csahin2007afford}
Erol {\c{S}}ahin, Maya {\c{C}}akmak, Mehmet~R Do{\u{g}}ar, Emre U{\u{g}}ur, and
  G{\"o}kt{\"u}rk {\"U}{\c{c}}oluk.
\newblock To afford or not to afford: A new formalization of affordances toward
  affordance-based robot control.
\newblock {\em Adaptive Behavior}, 15(4):447--472, 2007.

\bibitem[\protect\citeauthoryear{Saputra \bgroup \em et al.\egroup
  }{2019}]{saputra2019dynamic}
Azhar~Aulia Saputra, Wei~Hong Chin, Yuichiro Toda, Naoyuki Takesue, and Naoyuki
  Kubota.
\newblock Dynamic density topological structure generation for real-time ladder
  affordance detection.
\newblock In {\em IEEE/RSJ International Conference on Intelligent Robots and
  Systems}, pages 3439--3444. IEEE, 2019.

\bibitem[\protect\citeauthoryear{Saxena \bgroup \em et al.\egroup
  }{2014}]{saxena2014robobrain}
Ashutosh Saxena, Ashesh Jain, Ozan Sener, Aditya Jami, Dipendra~K Misra, and
  Hema~S Koppula.
\newblock Robobrain: Large-scale knowledge engine for robots.
\newblock {\em arXiv preprint arXiv:1412.0691}, 2014.

\bibitem[\protect\citeauthoryear{Seker \bgroup \em et al.\egroup
  }{2019}]{seker2019deep}
M~Yunus Seker, Ahmet~E Tekden, and Emre Ugur.
\newblock Deep effect trajectory prediction in robot manipulation.
\newblock {\em Robotics and Autonomous Systems}, 119:173--184, 2019.

\bibitem[\protect\citeauthoryear{Shu \bgroup \em et al.\egroup
  }{2016}]{shu2016learning}
Tianmin Shu, M.~S. Ryoo, and Song-Chun Zhu.
\newblock Learning social affordance for human-robot interaction.
\newblock In {\em International Joint Conference on Artificial Intelligence},
  pages 3454--3461, 2016.

\bibitem[\protect\citeauthoryear{Shu \bgroup \em et al.\egroup
  }{2017}]{shu2017learning}
Tianmin Shu, Xiaofeng Gao, Michael~S Ryoo, and Song-Chun Zhu.
\newblock Learning social affordance grammar from videos: Transferring human
  interactions to human-robot interactions.
\newblock In {\em IEEE International Conference on Robotics and Automation},
  pages 1669--1676. IEEE, 2017.

\bibitem[\protect\citeauthoryear{Sinapov and
  Stoytchev}{2007}]{sinapov2007learning}
Jivko Sinapov and Alexander Stoytchev.
\newblock Learning and generalization of behavior-grounded tool affordances.
\newblock In {\em IEEE 6th International Conference on Development and
  Learning}, pages 19--24. IEEE, 2007.

\bibitem[\protect\citeauthoryear{Song \bgroup \em et al.\egroup
  }{2010}]{song2010learning}
Dan Song, Kai Huebner, Ville Kyrki, and Danica Kragic.
\newblock Learning task constraints for robot grasping using graphical models.
\newblock In {\em IEEE/RSJ International Conference on Intelligent Robots and
  Systems}, pages 1579--1585. IEEE, 2010.

\bibitem[\protect\citeauthoryear{Song \bgroup \em et al.\egroup
  }{2013}]{song2013predicting}
Dan Song, Nikolaos Kyriazis, Iason Oikonomidis, Chavdar Papazov, Antonis
  Argyros, Darius Burschka, and Danica Kragic.
\newblock Predicting human intention in visual observations of hand/object
  interactions.
\newblock In {\em IEEE International Conference on Robotics and Automation},
  pages 1608--1615. IEEE, 2013.

\bibitem[\protect\citeauthoryear{Song \bgroup \em et al.\egroup
  }{2015}]{song2015task}
Dan Song, Carl~Henrik Ek, Kai Huebner, and Danica Kragic.
\newblock Task-based robot grasp planning using probabilistic inference.
\newblock {\em IEEE Transactions on Robotics}, 31(3):546--561, 2015.

\bibitem[\protect\citeauthoryear{Stark \bgroup \em et al.\egroup
  }{2008}]{stark2008functional}
Michael Stark, Philipp Lies, Michael Zillich, Jeremy Wyatt, and Bernt Schiele.
\newblock Functional object class detection based on learned affordance cues.
\newblock In {\em International Conference on Computer Vision Systems}, pages
  435--444. Springer, 2008.

\bibitem[\protect\citeauthoryear{Stoytchev}{2008}]{stoytchev2008learning}
Alexander Stoytchev.
\newblock Learning the affordances of tools using a behavior-grounded approach.
\newblock In {\em Towards Affordance-Based Robot Control}, pages 140--158.
  Springer, 2008.

\bibitem[\protect\citeauthoryear{Sun \bgroup \em et al.\egroup
  }{2010}]{sun2010learning}
Jie Sun, Joshua~L Moore, Aaron Bobick, and James~M Rehg.
\newblock Learning visual object categories for robot affordance prediction.
\newblock {\em The International Journal of Robotics Research},
  29(2-3):174--197, 2010.

\bibitem[\protect\citeauthoryear{Sun \bgroup \em et al.\egroup
  }{2014}]{sun2014object}
Yu~Sun, Shaogang Ren, and Yun Lin.
\newblock Object-object interaction affordance learning.
\newblock {\em Robotics and Autonomous Systems}, 62(4):487--496, 2014.

\bibitem[\protect\citeauthoryear{Szedmak \bgroup \em et al.\egroup
  }{2014}]{szedmak2014knowledge}
Sandor Szedmak, Emre Ugur, and Justus Piater.
\newblock Knowledge propagation and relation learning for predicting action
  effects.
\newblock In {\em IEEE/RSJ International Conference on Intelligent Robots and
  Systems}, pages 623--629. IEEE, 2014.

\bibitem[\protect\citeauthoryear{Thermos \bgroup \em et al.\egroup
  }{2017}]{thermos2017deep}
Spyridon Thermos, Georgios~Th Papadopoulos, Petros Daras, and Gerasimos
  Potamianos.
\newblock Deep affordance-grounded sensorimotor object recognition.
\newblock In {\em IEEE Conference on Computer Vision and Pattern Recognition},
  pages 6167--6175, 2017.

\bibitem[\protect\citeauthoryear{Tikhanoff \bgroup \em et al.\egroup
  }{2013}]{tikhanoff2013exploring}
V~Tikhanoff, U~Pattacini, L~Natale, and G~Metta.
\newblock Exploring affordances and tool use on the i{C}ub.
\newblock In {\em IEEE-RAS International Conference on Humanoid Robots}, pages
  130--137. IEEE, 2013.

\bibitem[\protect\citeauthoryear{Ugur and Piater}{2015}]{ugur2015refining}
Emre Ugur and Justus Piater.
\newblock Refining discovered symbols with multi-step interaction experience.
\newblock In {\em IEEE-RAS International Conference on Humanoid Robots}, pages
  1007--1012. IEEE, 2015.

\bibitem[\protect\citeauthoryear{Ugur \bgroup \em et al.\egroup
  }{2009}]{ugur2009affordance}
Emre Ugur, Erol Sahin, and Erhan Oztop.
\newblock Affordance learning from range data for multi-step planning.
\newblock In {\em International Conference on Epigenetic Robotics}, 2009.

\bibitem[\protect\citeauthoryear{Ugur \bgroup \em et al.\egroup
  }{2011}]{ugur2011goal}
Emre Ugur, Erhan Oztop, and Erol Sahin.
\newblock Goal emulation and planning in perceptual space using learned
  affordances.
\newblock {\em Robotics and Autonomous Systems}, 59(7-8):580--595, 2011.

\bibitem[\protect\citeauthoryear{Ugur \bgroup \em et al.\egroup
  }{2015}]{ugur2015staged}
Emre Ugur, Yukie Nagai, Erol Sahin, and Erhan Oztop.
\newblock Staged development of robot skills: Behavior formation, affordance
  learning and imitation with motionese.
\newblock {\em IEEE Transactions on Autonomous Mental Development},
  7(2):119--139, 2015.

\bibitem[\protect\citeauthoryear{Varadarajan and
  Vincze}{2012}]{varadarajan2012afrob}
Karthik~Mahesh Varadarajan and Markus Vincze.
\newblock Afrob: The affordance network ontology for robots.
\newblock In {\em IEEE/RSJ International Conference on Intelligent Robots and
  Systems}, pages 1343--1350. IEEE, 2012.

\bibitem[\protect\citeauthoryear{Veres \bgroup \em et al.\egroup
  }{2020}]{veres2020incorporating}
Matthew Veres, Ian Cabral, and Medhat Moussa.
\newblock Incorporating object intrinsic features within deep grasp affordance
  prediction.
\newblock {\em IEEE Robotics and Automation Letters}, 5(4):6009--6016, 2020.

\bibitem[\protect\citeauthoryear{Wang \bgroup \em et al.\egroup
  }{2013}]{wang2013robot}
Chang Wang, Koen~V Hindriks, and Robert Babuska.
\newblock Robot learning and use of affordances in goal-directed tasks.
\newblock In {\em IEEE/RSJ International Conference on Intelligent Robots and
  Systems}, pages 2288--2294. IEEE, 2013.

\bibitem[\protect\citeauthoryear{Ye \bgroup \em et al.\egroup
  }{2017}]{ye2017can}
Chengxi Ye, Yezhou Yang, Ren Mao, Cornelia Ferm{\"u}ller, and Yiannis
  Aloimonos.
\newblock What can {I} do around here? {D}eep functional scene understanding
  for cognitive robots.
\newblock In {\em IEEE International Conference on Robotics and Automation},
  pages 4604--4611. IEEE, 2017.

\bibitem[\protect\citeauthoryear{Zech \bgroup \em et al.\egroup
  }{2017}]{zech2017computational}
Philipp Zech, Simon Haller, Safoura~Rezapour Lakani, Barry Ridge, Emre Ugur,
  and Justus Piater.
\newblock Computational models of affordance in robotics: a taxonomy and
  systematic classification.
\newblock {\em Adaptive Behavior}, 25(5):235--271, 2017.

\bibitem[\protect\citeauthoryear{Zhu \bgroup \em et al.\egroup
  }{2014}]{zhu2014reasoning}
Yuke Zhu, Alireza Fathi, and Li~Fei-Fei.
\newblock Reasoning about object affordances in a knowledge base
  representation.
\newblock In {\em European Conference on Computer Vision}, pages 408--424.
  Springer, 2014.

\end{thebibliography}
}
\end{document}